%% file: root.tex
\documentclass[10pt,twocolumn,letterpaper]{article}
\usepackage{iccv}
\usepackage{times}
\usepackage{epsfig}
\usepackage{graphicx}
\usepackage{amsmath}
\usepackage{amssymb}
\usepackage{multirow}
\usepackage{relsize}
\usepackage{subfigure}
\usepackage{array}
\usepackage{color}

\usepackage[pagebackref=true,breaklinks=true,letterpaper=true,colorlinks,bookmarks=false]{hyperref}

\iccvfinalcopy % *** Uncomment this line for the final submission

\ificcvfinal\fi

\begin{document}

%%%%%%%%% TITLE
\title{Graspness Discovery in Clutters for Fast and Accurate Grasp Detection
}

\author{Chenxi Wang$^*$, Hao-Shu Fang$^*$, Minghao Gou, Hongjie Fang, Jin Gao, Cewu Lu$^\dag$\\
Shanghai Jiao Tong University\\
{\tt\small wcx1997@sjtu.edu.cn, fhaoshu@gmail.com, \{gmh2015, galaxies\}@sjtu.edu.cn, }\\
{\tt\small gao3944677@126.com, lucewu@sjtu.edu.cn}
}

\maketitle
% Remove page # from the first page of camera-ready.
\ificcvfinal\thispagestyle{empty}\fi

\newcommand\blfootnote[1]{%
  \begingroup
  \renewcommand\thefootnote{}\footnote{#1}%
  \addtocounter{footnote}{-1}%
  \endgroup
}

\blfootnote{$^*$ denotes equal contribution.\\ \indent \ Cewu Lu is the corresponding author, member of Qing Yuan Research
Institute and MoE Key Lab of Artificial Intelligence, AI Institute, Shanghai Jiao Tong University, China and Shanghai Qi Zhi institute.}

%%%%%%%%% ABSTRACT
\begin{abstract}
   Efficient and robust grasp pose detection is vital for robotic  manipulation. For general 6 DoF grasping, conventional methods treat all points in a scene equally and usually adopt uniform sampling to select grasp candidates. However, we discover that ignoring \textbf{where to grasp} greatly harms the speed and accuracy of current grasp pose detection methods. In this paper, we propose ``\textbf{graspness}'', a quality based on geometry cues that distinguishes graspable areas in cluttered scenes. A look-ahead searching method is proposed for measuring the graspness and statistical results justify the rationality of our method. To quickly detect graspness in practice, we develop a neural network named cascaded graspness model to approximate the searching process. Extensive experiments verify the stability, generality and effectiveness of our graspness model, allowing it to be used as a plug-and-play module for different methods. A large improvement in accuracy is witnessed for various previous methods after equipping our graspness model. Moreover, we develop GSNet, an end-to-end network that incorporates our graspness model for early filtering of low-quality predictions. Experiments on a large-scale benchmark, GraspNet-1Billion, show that our method outperforms previous arts by a large margin (\textbf{30+ AP}) and achieves a high inference speed. The library of GSNet has been integrated into \href{https://github.com/graspnet/anygrasp_sdk}{AnyGrasp}.
\end{abstract}

%%%%%%%%% BODY TEXT

\input{1_introduction}
\input{2_related_work}
\input{3_method}
\input{4_experiments}
\input{5_conclusion}

{\small
\bibliographystyle{ieee_fullname}
\bibliography{ref}
}

\input{6_appendix}

\end{document}

%% file: 1_introduction.tex
\vspace{-0.15in}
\section{Introduction}
\vspace{-0.05in}
As a fundamental problem in robotics, robust grasp pose detection for unstructured environment has been fascinating our community for decades. It has a broad spectrum of applications in picking~\cite{correll2016analysis}, assembling~\cite{suarez2018can}, home serving~\cite{edmonds2017feeling}, etc. Advancing the generality, accuracy and efficiency is a long pursuit of researchers in this field.
\input{charts/fig_introduction.tex}

For grasp pose detection in the wild, it can be regarded as a two-stage problem: given a single-view point cloud, we first find locations with high graspability (\emph{where} stage) and then decide grasp parameters like in-plane rotation, approaching depth, grasp score and gripper width (\emph{how} stage) for a local region. 

Previous methods for 6-DoF grasp pose detection in cluttered scenes mainly focused on improving the quality of grasp parameter prediction, \emph{i.e.}, the \emph{how} stage, and two lines of research are explored. The first line~\cite{gpd, pointnetgpd,6dofgraspnet} adopts a sampling-evaluation method, where grasp candidates are uniformly randomly sampled from the scene and evaluated by their model. The second line~\cite{s4g,graspnet,pointnetppgrasping} proposes end-to-end networks to calculate grasp parameters for the whole scene, where point clouds are sampled before~\cite{pointnetppgrasping} or during~\cite{s4g,graspnet} the forward propagation. For all these methods, the \emph{where} stage is not explicitly modeled (\textit{i.e.}, they do not perform a filtering procedure in a first stage) and candidate grasp points distribute uniformly in the scene.

However, we find that such uniform sampling strategy greatly hinders the performance of the whole pipeline. There are tremendous points in 3D contiguous space, while positive samples are concentrated in small local regions. Take GraspNet-1Billion~\cite{graspnet}, the current largest dataset in grasp pose detection as an example. We statistically find that, \emph{even with} object masks,  the graspable points are less than 10\% among all the samples, not to mention the candidate points in the whole scene. Such an imbalance causes a large waste of computing resources and degrades the efficiency. 

To tackle the above bottleneck in grasp pose detection, we propose a novel geometrically based quality, \emph{\textbf{graspness}}, for distinguishing graspable area in cluttered scenes. One might think that we need complex geometric reasoning to obtain such graspness. However, we discover that a simple look-ahead search by exhaustively evaluating possible future grasp poses from a point can well represent its graspness. Statistical results demonstrate the justifiability of our proposed graspness measure, where the local geometry around points with high graspness are distinguished from those with low scores.  Fig.~\ref{fig:introduction} gives an illustration of our graspness for a cluttered scene.

Furthermore, we develop a graspness model that approximates the above process in practice. Given a point cloud input, it predicts point-wise graspness score, which is referred to as \emph{graspable landscape}. 
Benefiting from the stability of the local geometry structures, our graspness model is object agnostic and robust to variation of viewpoint, scene, sensor, etc., making it a general and transferable module for grasp point sampling. We qualitatively evaluate its robustness and transferability in our analysis. Tremendous improvements in both speed and accuracy for previous sampling-evaluation based methods are witnessed after equipping them with our graspness model.

Based on our graspness model, we also propose Graspness-based Sampling Network (GSNet), an end-to-end two-stage network with a graspness-based sampling strategy. Our network takes a dense scene point cloud as input, which preserves the local geometry cues. The sampling layer firstly selects the points with high graspness. Remaining points are discarded from the forward propagation to improve the computation efficiency. Such two-stage design is beneficial to network convergence and also the final accuracy by providing more positive samples during training. 

We conduct extensive experiments to evaluate the effectiveness of our proposed graspness measure, model and the end-to-end network. Several baseline methods equipped with our graspness model outperform their vanilla counterparts by a large margin in both speed and accuracy. Moreover, our GSNet outperforms previous methods to a large extent. Our library has been integrated into AnyGrasp~\cite{fang2023anygrasp} to facilitate research in the robotics community.

%% file: charts/fig_introduction.tex
\begin{figure}[t]
    \centering
    \includegraphics[width=0.9\linewidth]{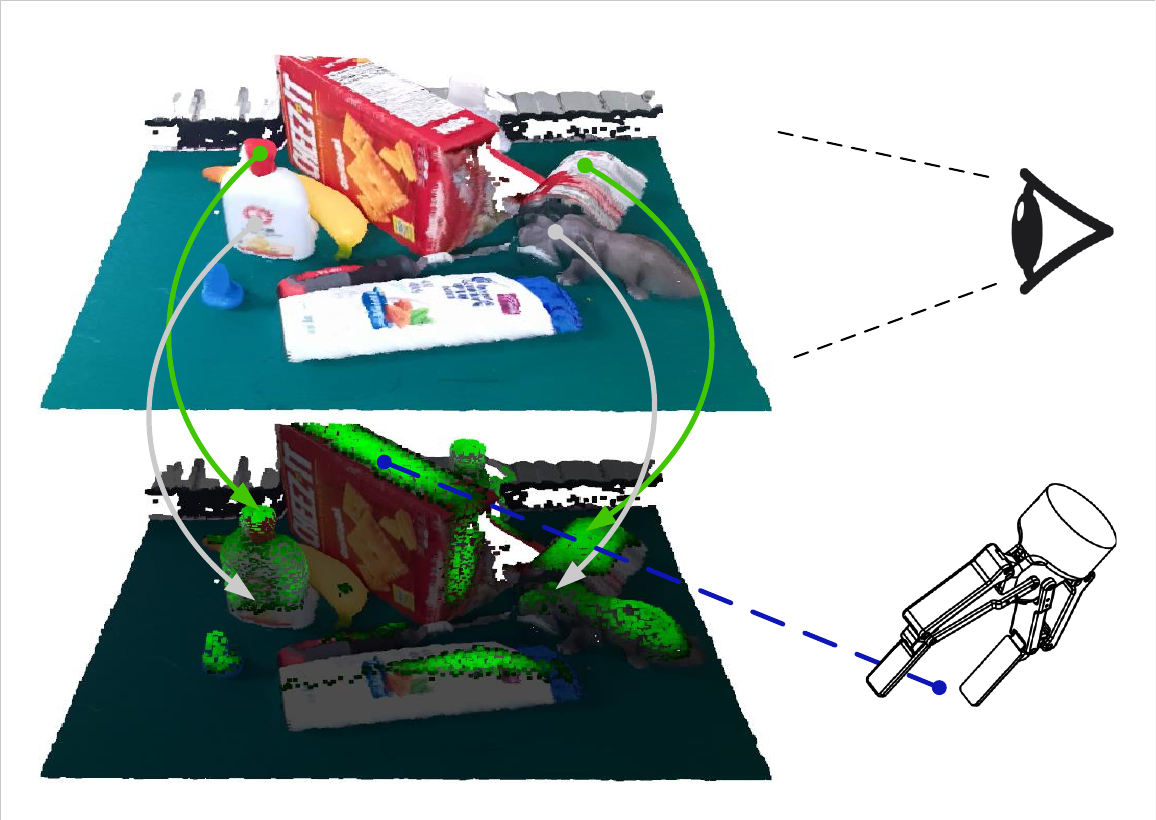}
    \caption{Graspness illustration for a cluttered scene. Brighter color denotes higher graspness. We prefer the points with high graspness for grasping.}
    \label{fig:introduction}
    \vspace{-0.2in}
\end{figure}

%% file: 2_related_work.tex
\section{Related Work}
In this section, we first briefly review previous methods on grasping in cluttered scenes, followed by concluding the common strategies they have used to sample grasp candidates. Finally we surveyed some literature in cognitive science area where graspness recognition is witnessed in human perception.
\vspace{-0.1in}
\paragraph{Grasping in Cluttered Scenes}
For cluttered scene grasp pose detection, previous research can be mainly divided into two categories: plannar based grasp detection and 6-DoF based grasp detection. The research in the first category~\cite{dsgd, cornell, lenz,dexnet2,ggcnn,2014redmon,kumra,multi, gou2021rgb, cao2021suctionnet} mainly took RGB images or depth images as inputs and output a set of rotated bounding boxes to represent the grasp poses. Due to the limitation of low DoF, their applications were usually restricted. Another line of research aimed to predict full DoF grasp poses. Among them, two different directions were explored. The first direction~\cite{varley2015generating,gpd,pointnetgpd,6dofgraspnet} adopted the sampling-evaluation based two-step policy, where grasp candidates were densely uniformly sampled in the scene and evaluated using a deep quality model. The second direction~\cite{graspnet,s4g,pointnetppgrasping,breyer2021volumetric} adopted the end-to-end strategy, where point clouds of the scene were directly processed by end-to-end networks. For each input point, the network attempted to predict the most feasible grasp pose. All the mentioned methods focused on improving the quality of grasp parameters, and the problem of where to grasp was not investigated.

\vspace{-0.1in}

\paragraph{Grasp Sampling Strategies} 
Several kinds of sampling strategies can be concluded from the above methods. The most common used strategy is the uniform sampling, which is adopted by~\cite{gpd,pointnetgpd,pointnetppgrasping,s4g}. Specifically, GPD~\cite{gpd} and PointNetGPD~\cite{pointnetgpd} uniformly sampled grasp points in the scene point cloud and estimated the rotation by darboux frame. Some end-to-end models~\cite{s4g, pointnetppgrasping} down-sampled the input point cloud by voxel grid to avoid memory explosion. A similar strategy, farthest point sampling, is adopted by other end-to-end model~\cite{graspnet}. Some optimization based methods are also explored. Ciocarlie \etal~\cite{ciocarlie2007dimensionality} and Hang \etal~\cite{hang2016hierarchical} adopted the simulated annealing method, while Mahler \etal~\cite{dexnet2} proposed cross-entropy methods. In~\cite{6dofgraspnet}, a grasp sampler network first sampled possible grasp poses on partial object point cloud and conducted iterative refinement by a grasp evaluator based on its gradient. In a recent paper by Clemens \etal~\cite{eppner2019billion}, several sampling methods for grasp dataset generation are reviewed. However, all the previous methods ignore the geometric cues for graspable point sampling. In this paper, we propose a novel graspness measure based on local geometry for graspable point sampling, which is much more efficient than previous uniform sampling and optimization based methods.

\vspace{-0.1in}

\paragraph{Graspness in Cognitive Science}
In cognitive area, researchers have studied the visual attention during grasping for a long period. Many literature \cite{baldauf2010attentional, bamford2020faster, gomez2018graspable, hannus2005selection, reed2010grab} demonstrated that human bias the allocation of available perceptual resources, named as affordance attention, towards the region with the highest graspability. And such attention usually precedes the action preparation stage~\cite{baldauf2010attentional}. Such discovery corresponds to our graspness concept and motivates us to apply it in the grasp sampling strategy.

%% file: 3_method.tex
\section{Graspness Discovery}
\subsection{Preliminary}
As mentioned above, we decouple the grasp pose detection problem into two stages. Before the common practice in previous research that directly calculates the grasp parameters, we first sample points and views with high graspness. Computational resources will be allocated to these areas thereafter to improve computational efficiency.

To determine the suitable grasp locations and the feasible approach directions with high graspability, we define two kinds of graspness in a high dimensional space to represent parallel attention in point locations and approach directions. Before detailing our graspness measure, we first introduce some basic notations.

For a point sets $\mathcal{P}=\{p_i| i=1,...,N\}$, we assume $V$ approach directions uniformly distributed in a sphere space $\mathcal{V}=\{v_j| j=1,...,V\}$.  

Two kinds of graspness scores are discussed in this paper. The first is the point-wise graspness scores denoted as 
$$\mathcal{S}^p=\{s^p_i| s^p_i \subset [0,1], i=1,...,N\},$$
where $[0,1]$ denotes that our graspness for each point ranges from 0 to 1. The second is the view-wise graspness scores denoted as
$$\mathcal{S}^v=\{s^v_i| s^v_i \subset [0,1]^V, i=1,...,N\},$$
where $[0,1]^V$ denotes $V$-dim graspness ranging in [0,1]. 

In the following section, we illustrate how we measure graspness for both single object and the  cluttered scene.

\subsection{Graspness Measure}
\label{sec:graspness}
\paragraph{Single Object Graspness}
Given an object point cloud, we aim to generate graspness for each point where higher activation denotes larger possibility for successful grasping. Assuming there is an oracle $\mathbf{1}(\cdot)$ that tells whether an arbitrary grasp is successful, and $\mathbf{G}_{i,j}$ denotes the set of all feasible grasp poses for view $v_j$ centered at point $p_i$, then the graspness score $\tilde{s}^\mathrm{p}_{i}$ and $\tilde{s}^\mathrm{v}_{i}$ can be obtained by an exhaustive look-ahead search:
\small
\begin{equation}
\label{eq:search}
\begin{split}
    \tilde{s}^\mathrm{p}_{i} = 
    & \frac{\sum_{j=1}^V\sum_{g\in\mathbf{G}_{i,j} }\mathbf{1}(g)}{\sum_{j=1}^V|\mathbf{G}_{i,j}|}, i=1,...,N,\\
    \tilde{s}^\mathrm{v}_{i} = 
    & \Big\{ \frac{\sum_{g\in\mathbf{G}_{i,j} }\mathbf{1}(g)}{|\mathbf{G}_{i,j}|} \Big| 1\le j \le V\Big\}, i=1,...,N.
\end{split}
\end{equation}
 \normalsize
By doing so, we guarantee that higher graspness value always denote higher possibility of successful grasping.

In practice, such an oracle $\mathbf{1}(\cdot)$ does not exist, and  $\mathbf{G}_{i,j}$ can contain infinite grasp poses in a continuous space. Thus, we make an approximation to the above process. For view $v_{j}$ of point $p_i$, we generate $L$ grasp candidates $\mathcal{G}_{i,j}=\{g^{i,j}_k|k=1,...,L\}$ by grid sampling along gripper depths and in-plane rotation angles. For each grasp $g^{i,j}_k$, we calculate a grasp quality score $q^{i,j}_k$ using a force analytic model~\cite{dexnet2}. A threshold $c$ is manually set to filter out unsuccessful grasps. Then, the relaxation form of Eqn.~\ref{eq:search} is:
\small
\begin{equation}
\label{eq:relax}
\begin{split}
    \tilde{s}^\mathrm{p}_{i} = 
    & \frac{\sum_{j=1}^V\sum_{k=1 }^L\mathbf{1}(q^{i,j}_k > c)}{\sum_{j=1}^V|\mathcal{G}_{i,j}|}, i=1,...,N,\\
    \tilde{s}^\mathrm{v}_{i} = 
    & \Big\{ \frac{\sum_{k=1}^L\mathbf{1}(q^{i,j}_k > c)}{|\mathcal{G}_{i,j}|} \Big| 1\le j \le V\Big\}, i=1,...,N.
\end{split}
\end{equation}
 \normalsize

\paragraph{Scene-Level Graspness}\label{sec:slg}
\input{charts/fig_graspness_collision}

After defining the object-level graspness, we extend it to cluttered scenes by first discussing the gap between them and then redefining the graspness in cluttered scenes.

A cluttered scene contains multiple objects and the irrelevant background. As shown in Fig.~\ref{fig:map_wo_colli}, the simplest way to compute scene-level graspness is directly projecting the object-level graspness score to the scene by object 6D poses. However, this solution ignores the differences between an object model and a scene cloud captured from RGB-D camera. Firstly, a valid grasp of a single object may collide with background or other objects when placing in cluttered manner and becomes a negative grasp. Secondly, as the depth camera provides single-view partial point clouds, we need to associate the scene point cloud with the projected object point.

To deal with the collision problem, we follow~\cite{graspnet} to reconstruct the scene using object 3D models and corresponding 6D poses.
Each grasp $g^{i,j}_k$ is evaluated by a collision checking process and assigned a collision label $c^{i,j}_k$. Our graspness scores are then updated as:
\small
\begin{equation}
\label{eq:scene}
\begin{split}
    \tilde{s}^\mathrm{p}_{i} = 
    & \frac{\sum_{j=1}^V\sum_{k=1 }^L\mathbf{1}(q^{i,j}_k > c)\cdot\mathbf{1}(c^{i,j}_k)}{\sum_{j=1}^V|\mathcal{G}_{i,j}|}, i=1,...,N,\\
    \tilde{s}^\mathrm{v}_{i} = 
    & \Big\{ \frac{\sum_{k=1}^L\mathbf{1}(q^{i,j}_k > c)\cdot\mathbf{1}(c^{i,j}_k)}{|\mathcal{G}_{i,j}|} \Big| 1\le j \le V\Big\}, i=1,...,N.
\end{split}
\end{equation}
 \normalsize
After that, we project the object points to the scene by object 6D poses. For each point in the scene, we obtain its graspness scores by nearest neighbor search and associate it with the nearest projected object point.

Finally, to obtain a coherent representation for the scene-level graspness scores, we perform a normalization for each scene:
\begin{equation}
\begin{split}
\mathcal{S}^p &=\Big\{\frac{\tilde{s}^p_i - \min(\tilde{\mathcal{S}}^p)}{\max(\tilde{\mathcal{S}}^p) - \min(\tilde{\mathcal{S}}^p)}\Big| i=1,...,N\Big\},\\
\mathcal{S}^v &=\Big\{\frac{\tilde{s}^v_i - \textbf{min}(\tilde{\mathcal{S}}^v)}{\textbf{max}(\tilde{\mathcal{S}}^v) - \textbf{min}(\tilde{\mathcal{S}}^v)}\Big| i=1,...,N\Big\},
\end{split}
\end{equation}
where $\textbf{min}(\cdot)$ denotes column wise minimum: 
\begin{equation}
\nonumber
\textbf{min}(\tilde{\mathcal{S}}^v) = \Big\{  \min\limits_{i=1}^N \tilde{s}_{i(j)}^v \Big| j=1,...,V
 \Big\},
\end{equation}
and so does $\textbf{max}(\cdot)$.
Fig.~\ref{fig:map_w_colli} shows an example of scene-level graspness scores.

\subsection{Justification}

In order to justify our graspness measure, we analyze the local geometry for regions with different graspness to find out whether they are really distinguishable geometrically. For a single-view point cloud, the cascaded graspness model detailed in Sec.~\ref{sec:cgm} is used to extract the local feature vector of each point. The points with graspness more than 0.3 are treated as positive samples, and negative ones of the same size are sampled with graspness less than 0.1. Fig.~\ref{fig:tsne} shows the t-SNE~\cite{tsne} visualization of the encoded local geometry (feature vectors of each point produced by backbone network) for all the scenes in GraspNet-1Billion~\cite{graspnet} training/testing set respectively. We can observe that regions with different graspness are quite distinguishable. It demonstrates that our graspness measure is rational and reveals the potential of learning graspness from point cloud.

\section{GSNet Architecture}
\vspace{-0.05in}
After defining the graspness measure, we introduce the end-to-end grasp pose detection network, GSNet, where our graspness is learned by an independent module and can be applied to other methods.
\input{charts/fig_tsne}
\input{charts/fig_architecture}

\subsection{Cascaded Graspness Model}\label{sec:cgm}
\vspace{-0.05in}
Given a dense single view point cloud $\mathcal{P}$, graspness model needs to learn two approximations: $f^p: \mathcal{P} \longrightarrow \mathcal{S}^p$ and $f^v: \mathcal{P} \longrightarrow \mathcal{S}^v$.

It is challenging to find a direct mapping from point coordinates to graspness scores due to the large domain gap between these two spaces. Instead, we decompose the whole process into two sub-functions. Consider a high dimensional feature set $\mathcal{F}$:
$$\mathcal{F} = \{f_i|f_i \subset \mathbb{R}^C, i=1,...,N\},$$
where $\mathbb{R}^C$ denotes $C$-dim feature space. The point set is firstly transformed to the feature set by $h^t: \mathcal{P}\longrightarrow\mathcal{F}$. Graspable landscapes are then generated by $h^p: \mathcal{F}\longrightarrow\mathcal{S}^p$ and $h^v: \mathcal{F}\longrightarrow\mathcal{S}^v$. Hence, we model the graspness scores by
$$f^p = h^p \circ h^t,\quad f^v = h^v \circ h^t,$$
where $\circ$ denotes function composition, and the feature set $\mathcal{F}$ is shared by both $h^p$ and $h^v$.

Although $h^p$ and $h^v$ can be learned simultaneously, the computation overhead is quite expensive since $\mathcal{S}^v$ is in high dimensional space. Meanwhile, it is not necessary to compute the view-wise graspable landscapes for all points since most of the points are not graspable at point level. Hence, we propose cascaded graspness model to learn $h^t$, $h^p$ and $h^v$ step by step, where points are sampled by the output of $h^p$ before learning $h^v$ to reduce computation cost.

\vspace{-0.15in}

\paragraph{Backbone Network}
Approximation of $h^t$ requires a strong backbone network for extraction of both global and local point features. We adopt ResUNet14 built upon MinkowskiEngine~\cite{minkowski} because it can flexibly process point sets of any size with sparse convolution and has shown excellent performance in multiple tasks of 3D deep learning~\cite{choy2019fully, gojcic2020learning, gwak2020generative, choy2020deep}. The network can also be replaced by other point-wise networks, such as PointNet~\cite{pointnet, pointnetpp}, PointCNN~\cite{li2018pointcnn} and SSCNs~\cite{graham20183d}.

The network adopts a U-shape architecture with residual blocks, which obtains point features using 3D sparse (transposed) convolutions and skip-connections. For a point cloud of size $N\times 3$, it extracts a $C$-channel feature vector set, and outputs a point set of size $N\times(3+C)$ for graspable sampling and grasp generation.

\vspace{-0.15in}

\paragraph{Graspable Farthest Point Sampling}
The modeling for $h^p$ is implemented with a multi-layer perceptron (MLP) network to generate point-wise graspable landscape. Specifically, the output contains a prediction for the graspable landscape of size $N\times 1$ and a binary objectness classification scores of size $N\times 2$, resulting a total output of size $N\times 3$. Graspness scores of non-object points are set to 0.

After obtaining the point-wise graspable landscape, we select points with graspness score larger than $\delta^\mathrm{p}$ and adopt farthest point sampling (FPS) to maximize distances among sampled points. $M$ seed points are sampled with $(3+C)$-dim features, where $3$ denotes the point coordinates and $C$ denotes the features output by the backbone network.

\vspace{-0.1in}

\paragraph{Graspable Probabilistic View Selection}
$h^v$ is also modeled by an MLP. We apply it to the sampled seed points and output $M\times V$ vectors for view-wise graspable landscapes and $M\times C$ residual features for grasp generation. $V$ views are sampled from a unit sphere using Fibonacci lattices~\cite{gonzalez2010measurement}.

After obtaining the view-wise graspness scores, we select the best view for afterward predictions during inference. For training, we adopt probabilistic view selection (PVS) that normalizes the graspness scores of all views on a seed point to (0,1) and regard them as probability scores, according to which the view is sampled. The $M$ seed point-view pairs are then used to estimate grasp scores, gripper widths, approach distances and in-plane rotation angles.

\subsection{Grasp Operation Model}
Crop-and-refine has been proven effective in estimating candidate configuration in both 2D and 3D tasks~\cite{fasterrcnn, maskrcnn, votenet}. We crop points in directional cylinder spaces which are generated by seed point-view pairs, transform them to gripper frames and estimate their grasp parameters.

\vspace{-0.1in}

\paragraph{Cylinder-Grouping from Seed Points}
The locations and directions of cylinder spaces are determined by seed point coordinates and view vectors respectively.
For each of the $M$ point-view pairs, we group and sample $K$ points from $M$ seed points using the cylinder with fixed height $d$ and radius $r$. After aligning the cylinder with gripper frame as~\cite{graspnet}, the point coordinates are normalized by cylinder radius and concatenated with feature vectors which are the sum of features output by graspable FPS and graspable PVS. The grouped point sets of size $M\times K\times (3+C)$ are called grasp candidates, where $K$ stands for the number of sampled points in each group.

\vspace{-0.1in}

\paragraph{Grasp Generation from Candidates} We use a shared PointNet~\cite{pointnet} for grasp generation. Grasp candidates are processed by an MLP network and a max-pooling layer, and be output as feature vectors of size $M\times C^{'}$. Finally we get grasp configurations by a new MLP network.

The output of GSNet contains scores and widths for different (in-plane rotation)-(approach depth) combinations. We pick the combination with the highest score as the grasp prediction. The output size is $M\times(A\times D\times 2)$, where $A$ denotes the number of in-plane rotation angles, $D$ denotes the number of gripper depths and $2$ denotes the score and the width.

\vspace{-0.1in}

\paragraph{Grasp Score Representation}
We use the minimum friction coefficient $\mu$ under which a grasp is antipodal to evaluate the quality of the grasp. Based on this, we define the grasp score as
\begin{equation}
    q_i =
    \begin{cases}
    \frac{\ln{(\mu_\mathrm{max} / \mu_i)}}{\ln{(\mu_\mathrm{max} / \mu_\mathrm{min})}} &\quad g_i\text{ is positive,}\\
    0 &\quad g_i\text{ is negative.}
    \end{cases}
\end{equation}

All scores are normalized to $[0,1]$. Smaller $\mu_i$ indicates higher score $q_i$ and more probability to succeed.

\subsubsection{Loss Function}
Cascaded graspness model and grasp operation model are trained simultaneously with multi-task losses:
\begin{equation}
    L = L_{o} + \alpha(L_{p} + \lambda L_{v}) +\\ \beta(L_{s} + L_{w}),
\end{equation}
where $L_{o}$ is for objectness classification, $L_{p}$, $L_{v}$, $L_{s}$ and $L_{w}$ are for regressions of point-wise graspable landscape, view-wise graspable landscape, grasp scores and gripper widths respectively. $L_{p}$ and $L_{s}$ are calculated only if the related points are on objects, $L_{v}$ is calculated for views on seed points and $L_{w}$ is calculated for grasp poses with ground truth scores $>$ 0. We use softmax for classification tasks and smooth-$l_1$ loss for regression tasks.

%% file: charts/fig_graspness_collision.tex
\begin{figure}[t]
    \centering
    \subfigure[without collision.]
    {
        \label{fig:map_wo_colli}
        \centering
        \includegraphics[width=0.40\columnwidth]{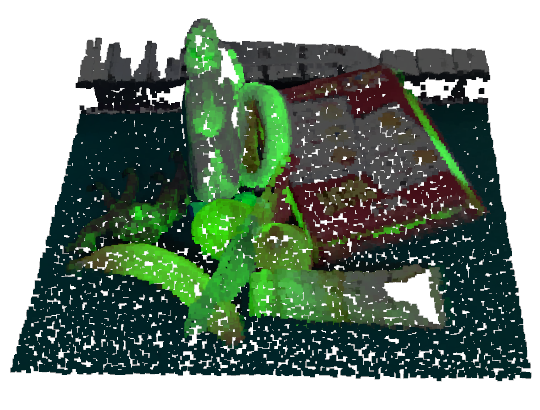}
    }
    \subfigure[with collision.]
    {
        \centering
        \includegraphics[width=0.40\columnwidth]{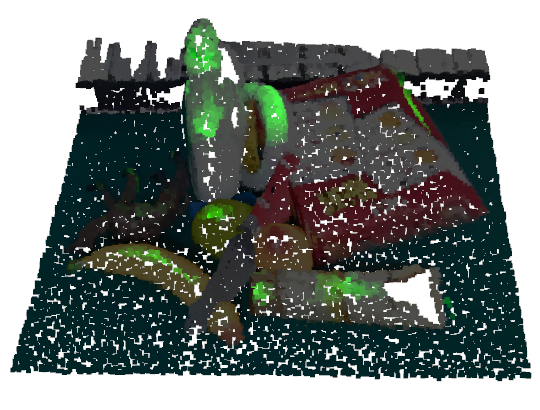}
        \label{fig:map_w_colli}
    }
    \caption{Graspness scores. The left image shows the graspness without collision detection while the right image shows the graspness with collision detection}
    \label{fig:graspness_vis}
\vspace{-0.2in}
\end{figure}

%% file: charts/fig_tsne.tex
\begin{figure}[t]
    \centering
    \subfigure[Training set.]
    {
        \centering
        \fbox{\includegraphics[width=0.4\linewidth]{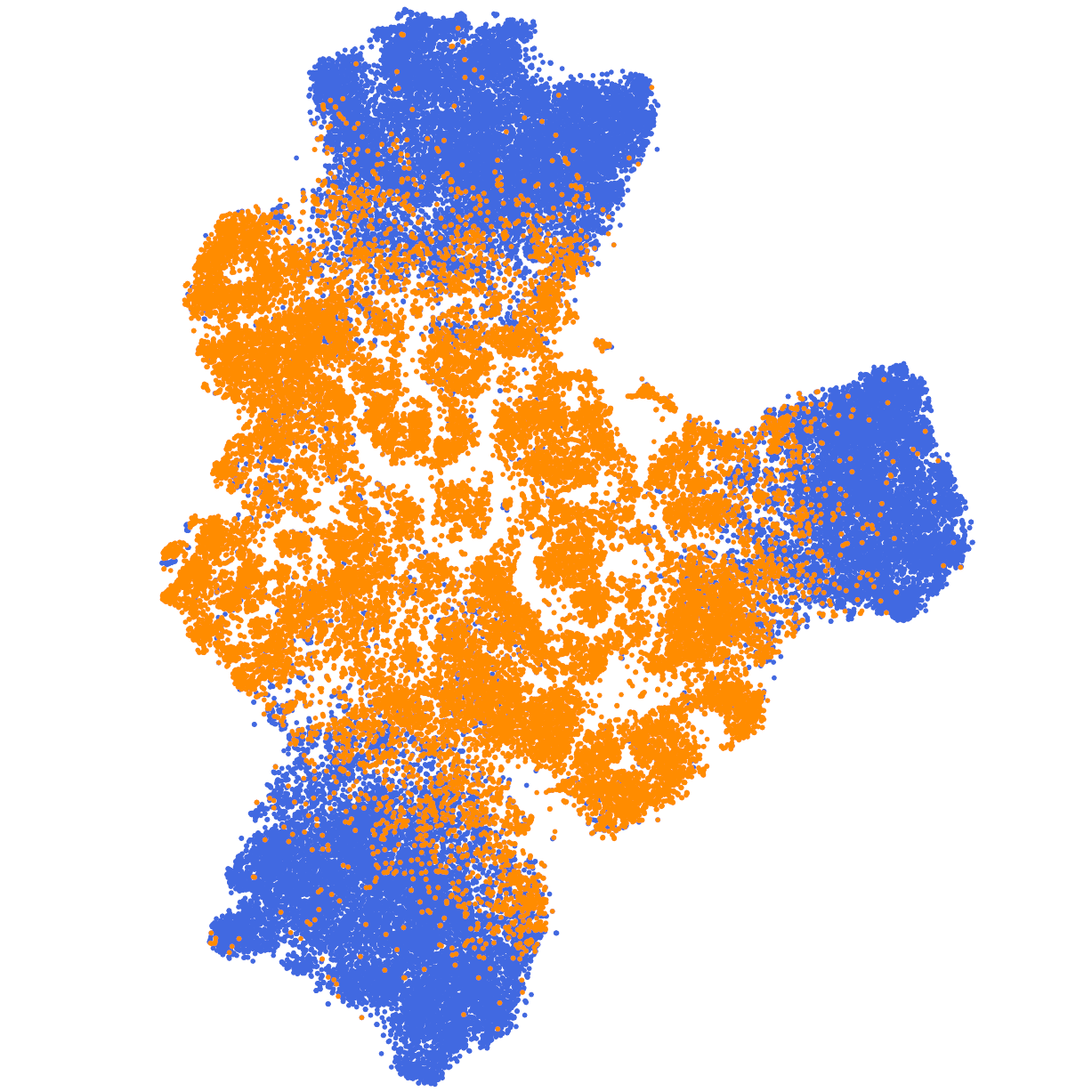}}
    }
    \subfigure[Testing set.]
    {
        \centering
        \fbox{\includegraphics[width=0.4\linewidth]{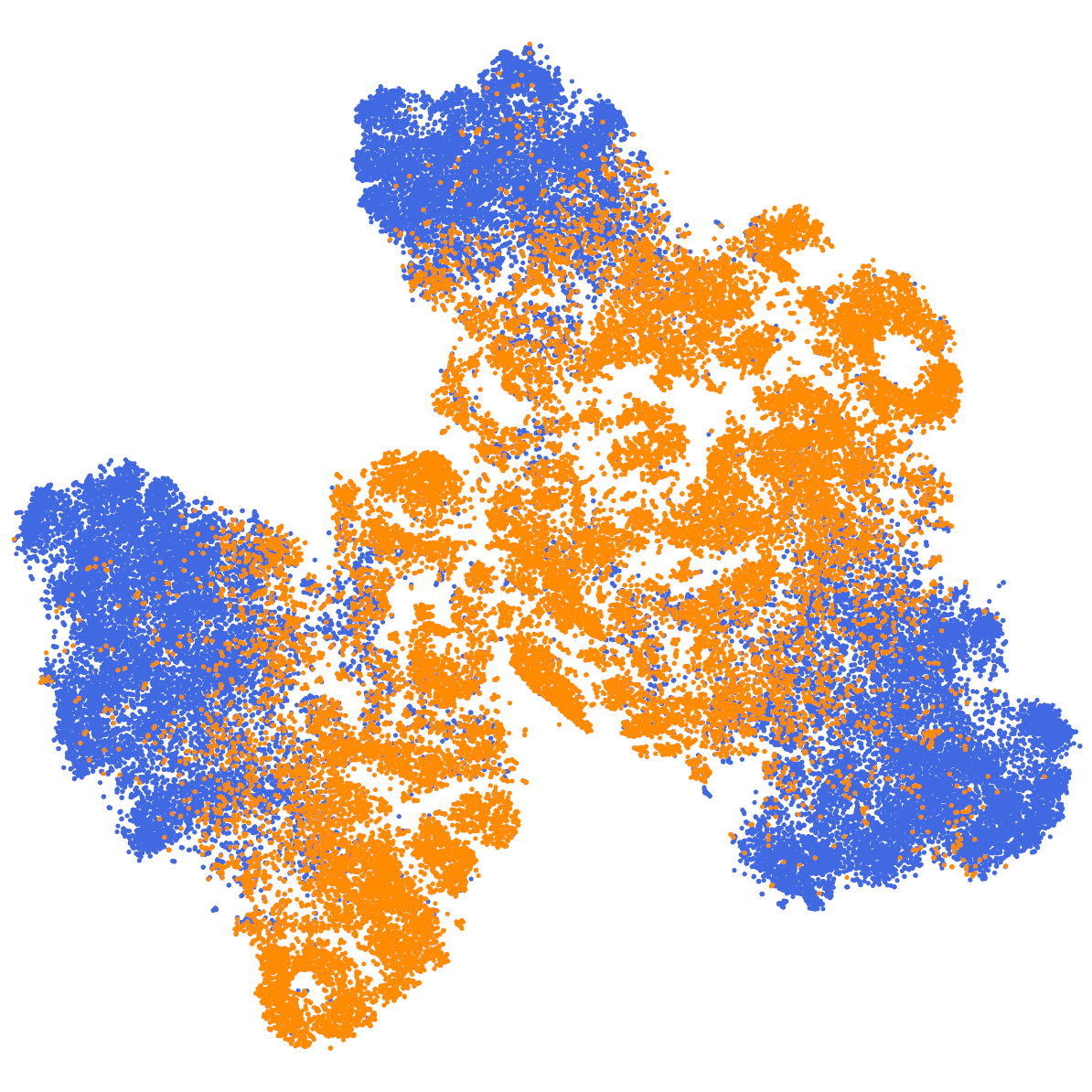}}
    }
    \caption{t-SNE visualization of encoded local geometry. Orange points denote the samples with high graspness, and blue points denote the  samples with low graspness.}
    \label{fig:tsne}
    \vspace{-0.2in}
\end{figure}

%% file: charts/fig_architecture.tex
\begin{figure*}
    \centering
    \includegraphics[width=0.9\linewidth]{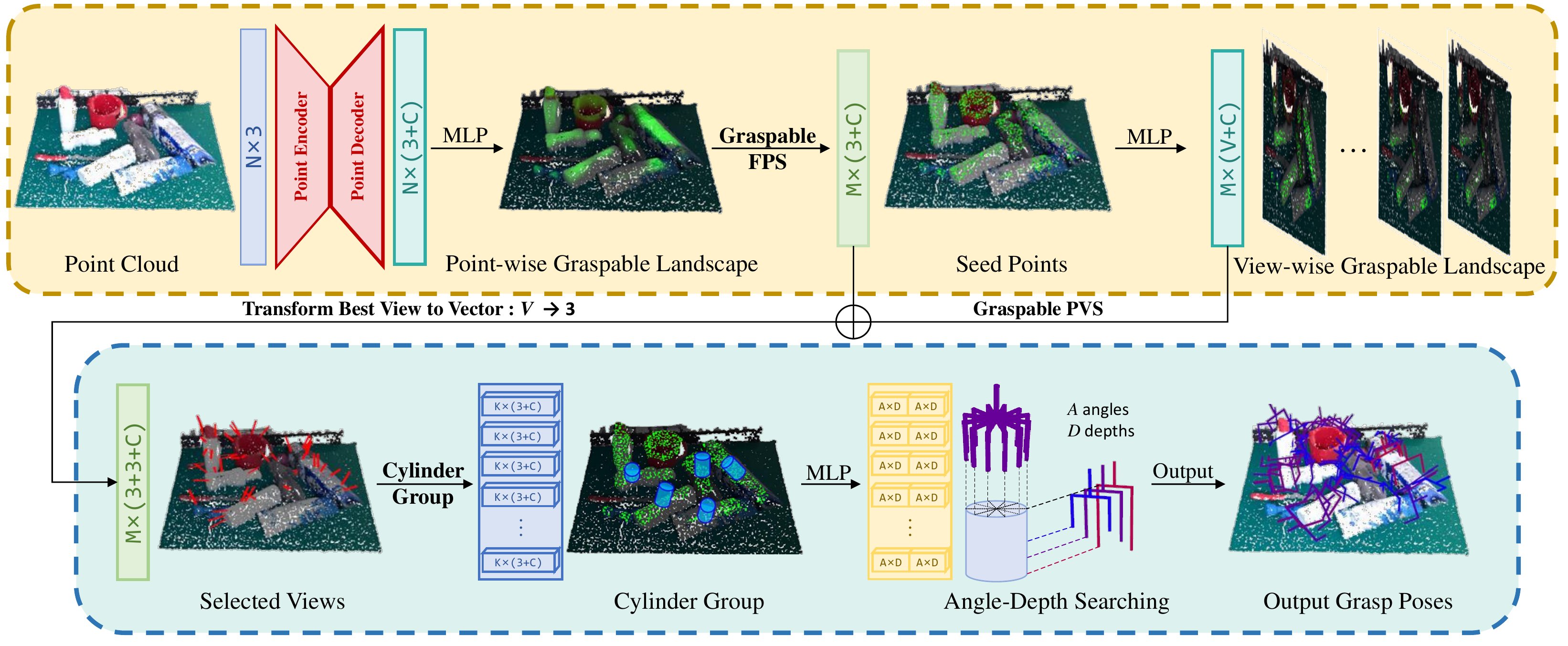}
    \caption{GSNet architecture. The two rows show the process of cascaded graspness model and grasp operation model respectively. In cascaded graspness model, point encoder-decoder outputs $C$-dim feature vectors for the input $N$ points. A point-wise graspable landscape is generated and $M$ seed points are sampled from it. The seeds are then used to generate view-wise graspable landscapes, and select the grasp view. In grasp operation model, the seeds are grouped in cylinder regions. The grasp scores and gripper widths are predicted for each group and used to output $M$ grasp poses.}
    \label{fig:architecture}
    \vspace{-0.1in}
\end{figure*}

%% file: 4_experiments.tex
\section{Experiments}

\subsection{Implementation Details}
\paragraph{Benckmark Dataset}
GraspNet-1Billion~\cite{graspnet} is a large-scale dataset for grasp pose detection, which contains 190 scenes with 256 different views captured by two cameras (RealSense/Kinect). The testing scenes are divided into three splits according to the object categories (seen/similar/novel). A unified evaluation metric is proposed to benchmark both image based methods and point cloud based methods. 
We adopt this benchmark as it aligns well with real-world grasping.

\vspace{-0.1in}

\paragraph{Data Processing and Augmentation}
The point cloud is downsampled with voxel size 0.005m before being fed into the network, and contains only XYZ in camera coordinates. Input clouds are augmented on the fly by random flipping along YZ plane and random rotation around Z axis in $\pm 30^\circ$.

\vspace{-0.1in}

\paragraph{Implementations}
To obtain graspness for scenes in GraspNet-1Billion, we follow the process illustrated in Sec.~\ref{sec:graspness} since it contains abundant grasp pose annotations. For each point, it densely labels grasp quality score for 300 different views and 48 grasps for each view. Thus, our approach directions $V$ and grasp candidates per view $L$ are set as 300 and 48.

For our network, the backbone network adopts an encoder-decoder architecture and outputs feature vectors of channel $C=512$. In graspable FPS/PVS, $M=1024$ seed points and $V=300$ views are sampled, and the threshold $\delta^\text{P}$ is set to 0.1. The size of MLP used for $h^p$ is $(512, 3)$ and  $h^v$ is $(512, 512, 300)$. In cylinder-grouping, we sample $K=16$ seed points in the cylinder space with radius $r=0.05$m and height range of $[-0.02\mathrm{m}, 0.04\mathrm{m}]$. We divide in-plane rotation angles into $A=12$ classes (15$^\circ$ per class) and use $D=4$ classes for approaching distances ($0.01$m, $0.02$m, $0.03$m, $0.04$m). The two MLPs used to process attentional proposals and output grasp scores and gripper widths have the size of $(512, 256, 256)$ and $(256, 256, 96)$ respectively. Finally the network outputs grasp scores and gripper widths for $A\times D=48$ classes. In loss functions, we set $\alpha, \beta, \lambda = 10, 10, 10$. 

\vspace{-0.1in}

\paragraph{Training and Inference}
Our model is implemented with PyTorch and trained on Nvidia GTX 1080Ti GPUs for 10 epochs with Adam optimizer~\cite{adam} and the batch size of 4. The learning rate is 0.001 at the first epoch, and multiplied by 0.95 every one epoch. The network takes about 1 day to converge. During training, we use one GPU for model update and one GPU for label generation. In inference, we only use one GPU for fast prediction.

\subsection{Performance of Cascaded Graspness Model}
\input{charts/tab_cgm_ablation}
\input{charts/tab_benchmark}

Cascaded graspness model is proposed to distinguish graspable areas in various scenes, thus the generality and stability across different domains are important for the model. Here we design an experiment to illustrate its generality and stability.

\paragraph{Evaluation Metric} The \textit{ranking error} is used to quantitatively evaluate the function approximation ability of the model. We divide the range of graspness score into $K$ bins uniformly and convert the contiguous scores to discrete ranks. The ranking error is defined as the mean rank distances between predictions and labels:
\begin{equation}
    e_\mathrm{rank} = \frac{1}{N_r}\sum_{i=1}^{N_r}\frac{|\hat{r}_i - r_i|}{K},
\end{equation}
where $r_i,\hat{r}_i\in\{0,1,...,K-1\}$ stand for the ranks for predictions and labels respectively, and $N_r$ is the number of predictions. We set $K=20$ in experiments. $e^p_\mathrm{rank}$ and $e^v_\mathrm{rank}$ are used to denote the ranking error of point-wise graspness score and view-wise graspness score respectively.

\paragraph{Inference in Different Domains}
We conduct three groups of experiments where the dataset is split by object categories, viewpoints and cameras respectively (detailed in Tab.~\ref{tab:cgm_ablation}). In the first group, we train the model on scene 0-99, and test it on scenes with three object categories (seen, similar and novel). The second group divides the 256 viewpoints into 3 sets, trains the model on viewpoint 0-127, and tests on three viewpoint sets respectively. The third group trains the model on Kinect captured data, and tests the performance on data captured by RealSense.

The results are summarized in Tab.~\ref{tab:cgm_ablation}. For point-wise graspness prediction, we can see that the difference between $e^p_\mathrm{rank}$ of seen and novel categories is not obvious. View variation also has a low impact on point-wise graspness prediction. The $e^p_\mathrm{rank}$ of RealSense is higher than Kinect, but the distance is still in an acceptable range. For view-wise graspness prediction, $e^v_\mathrm{rank}$ in all groups are nearly unchanged. These experiments prove the stability and generality of the cascaded graspness model when transferred to new domains.

\subsection{Comparing with Representative Methods}
\input{charts/fig_result}
We compare our method with previous representative methods. GG-CNN~\cite{ggcnn} and Chu~\etal~\cite{multi} are rectangle based methods which take images as input. GPD~\cite{gpd} and Liang~\etal~\cite{pointnetgpd} classify grasp candidates generated by rule-based point cloud sampling. Fang~\etal~\cite{graspnet} propose an end-to-end network which predicts grasp poses directly from scene point clouds.

We test our method in three object categories respectively and report the results in Tab.~\ref{tab:benchmark}. The models for RealSense and Kinect are trained separately. Our method outperforms previous methods by a large margin on both cameras without any post-processing. Compared with Fang~\etal, the previous state-of-the-art method, GSNet improves the performance by $\sim$2x on \textbf{AP} metric~\cite{graspnet}. Notably, on the most difficult metric $\text{AP}_\text{0.4}$, GSNet still achieves a great relative improvement ($>140\%$) on all categories. Fig.~\ref{fig:result} presents the qualitative results of our network. The top-1 grasp accuracy on three categories are 78.22/76.49, 62.88/57.64 and 28.97/24.04 for Realsense/Kinect input.

We also report the results after simple collision detection using a parallel-jaw gripper model, where all grasps collided with scene points are removed. The results are improved by 1.42/2.31 \textbf{AP}, 1.06/1.79 \textbf{AP} and 0.33/0.77 \textbf{AP} on the three categories respectively.

\vspace{-0.1in}

\subsection{Boosting with Cascaded Graspness Model}
We apply the cascaded graspness model (CGM) to GPD, Liang~\etal and Fang~\etal directly and compare the results with the original methods. For Fang~\etal, we simply replace ApproachNet with our module. For GPD and Liang~\etal, we first determine the grasp candidate points using our predicted point-wise graspable landscape, followed by their post processing of Darboux frame estimation and grasp images/clouds classification.

In the middle of Tab.~\ref{tab:benchmark}, we show the results after adding the CGM. Both the two-step methods and the end-to-end method achieve significant performance gains, proving the effectiveness of cascaded graspness model. Graspable landscapes can not only improve candidate qualities, but also reduce the huge computation time caused by densely sampling.
\\
\subsection{Analysis}

\paragraph{Effects of Graspable FPS/PVS}

In Sec.~\ref{sec:cgm} we use graspable FPS to sample seed points from graspable landscapes, while other sampling methods can also be applied to the network. We compare our sampling method with three alternatives: a) random sampling from the whole point cloud; b) FPS from the whole point cloud; c) random sampling from graspable landscapes. Tab.~\ref{tab:fps_pvs} shows the results of the models trained using different sampling methods. FPS outperforms random sampling by at least 4.98 \textbf{AP} and sampling with graspable landscapes improves the results by over 7 \textbf{AP} for both FPS and random sampling, which proves the effectiveness of graspable FPS.

For view selection, we compare graspable PVS with two methods: a) selecting views by surface normal; b) selecting the view with the highest graspness score during training. The results in Tab.~\ref{tab:fps_pvs} show that our method outperforms both alternative strategies. Graspable PVS dynamically selects approach vectors, which provides richer data for model training than other methods.

\paragraph{Selection of Landscape Representations}
\input{charts/tab_fps_pvs}
\input{charts/tab_graspness_selection}
In Sec.~\ref{sec:slg} we extend object-level graspness scores to cluttered scenes. Tab.~\ref{tab:point_graspness} shows that sampling from scene-level graspness performs better than object-level counterpart. The representation for graspness score also has multiple choices. We replace the original definition \emph{ratio of feasible grasps} with mean and maximum grasp quality scores respectively in the calculation of view-wise graspness scores, and the results in Tab.~\ref{tab:view_graspness} shows that feasible grasp ratio performs the best.

\paragraph{Model Speed}
\input{charts/tab_speed}

Tab.~\ref{tab:speed} shows the inference time of our method. Cascaded graspness model achieves a high speed on RealSense/Kinect data, which can also provide accurate sampling for various grasp detection methods. GPD and
PointNetGPD take $>$1s while ours takes only $\sim$0.1s.

\subsection{Real Grasping Experiments}
We also conduct grasping experiments for cluttered scenes in the real-world setting. The configuration of our experimental setup is illustrated in supplementary materials. The experiments are conducted on a UR-5 robotic arm with an Intel RealSense D435 camera and a Robotiq two-finger gripper. During experiments, we only keep the points on table workspace for speed up. 

\input{charts/tab_realgrasp_scene}
We conduct grasping experiments in six cluttered scenes. Each scene contains 6-8 objects selected from GraspNet-1Billion. Objects are put together randomly and we repeat the grasping pipeline until the table are cleaned. The success rate is defined as the ratio of object number and attempt number. Tab.~\ref{tab:realgrasp_scene} reports the grasping performance, which proves the effectiveness of our method. A comparison with other baselines is detailed in supplementary materials.

%% file: charts/tab_cgm_ablation.tex
\begin{table*}[h]
    \centering\small
    \begin{tabular}{c||c|ccc||c|ccc||c|cc}
        \hline
        & \multicolumn{4}{c||}{Object Variation} & \multicolumn{4}{c||}{Viewpoint Variation} & \multicolumn{3}{c}{Camera Variation}\\
        \cline{2-12}
        & Train & \multicolumn{3}{c||}{Test} & Train & \multicolumn{3}{c||}{Test} & Train & \multicolumn{2}{c}{Test} \\
        \hline
        Scene & 0-99 & 100-129 & 130-159 & 160-189 & 0-99 & 100-129 & 100-129 & 100-129 & 0-99 & 100-129 & 100-129 \\
        \hline
        View & 0-255 & 0-255 & 0-255 & 0-255 & 0-127 & 0-127 & 128-191 & 192-255 & 0-255 & 0-255 & 0-255 \\
        \hline
        Camera & Kinect & Kinect & Kinect & Kinect & Kinect & Kinect & Kinect & Kinect & Kinect & Kinect & Realsense \\
        \hline
        $e^p_\mathrm{rank}$ & 0.0485 & 0.0677 & 0.0856 & 0.0802 & 0.0484 & 0.0697 & 0.0725 & 0.0763 & 0.0485 & 0.0677 & 0.0984\\
        $e^v_\mathrm{rank}$ & 0.0451 & 0.0457 & 0.0459 & 0.0413 & 0.0458 & 0.0468 & 0.0473 & 0.0476 & 0.0451 & 0.0457 & 0.0461 \\
        \hline
    \end{tabular}
    \caption{Ranking error of cascaded graspness model on different test setting. We can see that the graspness model is not sensitive to object/viewpoint/camera variations.}
    \label{tab:cgm_ablation}
\end{table*}

%% file: charts/tab_benchmark.tex
\begin{table*}[]
    \centering
    \scalebox{0.9}{
    \begin{tabular}{c|ccc|ccc|ccc}
    \hline
    \multirow{2}{*}{{Methods}} & \multicolumn{3}{c|}{{Seen}} & \multicolumn{3}{c|}{{Similar}} & \multicolumn{3}{c}{{Novel}}\\
    \cline{2-10}
    & {\textbf{AP}} & {$\text{AP}_{0.8}$} & {$\text{AP}_{0.4}$} & {\textbf{AP}} & {$\text{AP}_{0.8}$} & {$\text{AP}_{0.4}$} & {\textbf{AP}} & {$\text{AP}_{0.8}$} & {$\text{AP}_{0.4}$}\\
    \hline
    \footnotesize{GG-CNN~\cite{ggcnn}} & \footnotesize{15.48/16.89} & \footnotesize{21.84/22.47} & \footnotesize{10.25/11.23} & \footnotesize{13.26/15.05} & \footnotesize{18.37/19.76} & \footnotesize{4.62/6.19} & \footnotesize{5.52/7.38} & \footnotesize{5.93/8.78} & \footnotesize{1.86/1.32}\\
    \footnotesize{Chu~\etal~\cite{multi}} & \footnotesize{15.97/17.59} & \footnotesize{23.66/24.67} & \footnotesize{10.80/12.74} & \footnotesize{15.41/17.36} & \footnotesize{20.21/21.64} & \footnotesize{7.06/8.86} & \footnotesize{7.64/8.04} & \footnotesize{8.69/9.34} & \footnotesize{2.52/1.76}\\
    \footnotesize{GPD~\cite{gpd}} & \footnotesize{22.87/24.38} & \footnotesize{28.53/30.16} & \footnotesize{12.84/13.46} & \footnotesize{21.33/23.18} & \footnotesize{27.83/28.64} & \footnotesize{9.64/11.32} & \footnotesize{8.24/9.58} & \footnotesize{8.89/10.14} & \footnotesize{2.67/3.16}\\
    \footnotesize{Liang~\etal~\cite{pointnetgpd}} & \footnotesize{25.96/27.59} & \footnotesize{33.01/34.21} & \footnotesize{15.37/17.83} & \footnotesize{22.68/24.38} & \footnotesize{29.15/30.84} & \footnotesize{10.76/12.83} & \footnotesize{9.23/10.66} & \footnotesize{9.89/11.24} & \footnotesize{2.74/3.21}\\
    \footnotesize{Fang~\etal~\cite{graspnet}} & \footnotesize{27.56/29.88} & \footnotesize{33.43/36.19} & \footnotesize{16.95/19.31} & \footnotesize{26.11/27.84} & \footnotesize{34.18/33.19} & \footnotesize{14.23/16.62} & \footnotesize{10.55/11.51} & \footnotesize{11.25/12.92} & \footnotesize{3.98/3.56}\\
    \hline
    \footnotesize{GPD + \textit{CGM}} & \footnotesize{28.16/29.65} & \footnotesize{34.07/35.59} & \footnotesize{17.21/18.94} & \footnotesize{26.47/28.19} & \footnotesize{33.14/33.74} & \footnotesize{14.27/16.20} & \footnotesize{9.73/10.89} & \footnotesize{10.55/11.37} & \footnotesize{3.35/4.12}\\
    \footnotesize{Liang~\etal + \textit{CGM}} & \footnotesize{33.86/33.17} & \footnotesize{41.50/40.85} & \footnotesize{22.93/23.18} & \footnotesize{28.91/29.06} & \footnotesize{34.70/35.96} & \footnotesize{16.95/17.33} & \footnotesize{11.97/12.47} & \footnotesize{13.52/13.31} & \footnotesize{4.01/4.64}\\
    \footnotesize{Fang~\etal + \textit{CGM}} & \footnotesize{41.46/39.51} & \footnotesize{49.32/48.75} & \footnotesize{29.64/26.19} & \footnotesize{36.87/35.28} & \footnotesize{45.69/44.93} & \footnotesize{25.29/23.84} & \footnotesize{15.11/13.26} & \footnotesize{17.49/15.03} & \footnotesize{6.74/5.28}\\
    \hline
    \footnotesize{Ours} & \footnotesize{65.70/61.19} & \footnotesize{76.25/71.46} & \footnotesize{\textbf{61.08}/56.04} & \footnotesize{53.75/47.39} & \footnotesize{65.04/56.78} & \footnotesize{45.97/40.43} & \footnotesize{23.98/19.01} & \footnotesize{29.93/23.73} & \footnotesize{14.05/10.60}\\
    \footnotesize{Ours + \textit{CD}} & \footnotesize{\textbf{67.12}/\textbf{63.50}} & \footnotesize{\textbf{78.46}/\textbf{74.54}} & \footnotesize{60.90/\textbf{58.11}} & \footnotesize{\textbf{54.81}/\textbf{49.18}} & \footnotesize{\textbf{66.72}/\textbf{59.27}} & \footnotesize{\textbf{46.17}/\textbf{41.89}} & \footnotesize{\textbf{24.31}/\textbf{19.78}} & \footnotesize{\textbf{30.52}/\textbf{24.60}} & \footnotesize{\textbf{14.23}/\textbf{11.17}}\\
    \hline
    \end{tabular}}
    \caption{GraspNet-1Billion evaluation results on RealSense/Kinect. \textit{CGM} is cascaded graspness model. \textit{CD} is collision detection.}
    \label{tab:benchmark}
    \vspace{-0.1in}
\end{table*}

%% file: charts/fig_result.tex
\begin{figure*}[h]
    \centering
    \includegraphics[width=0.9\linewidth]{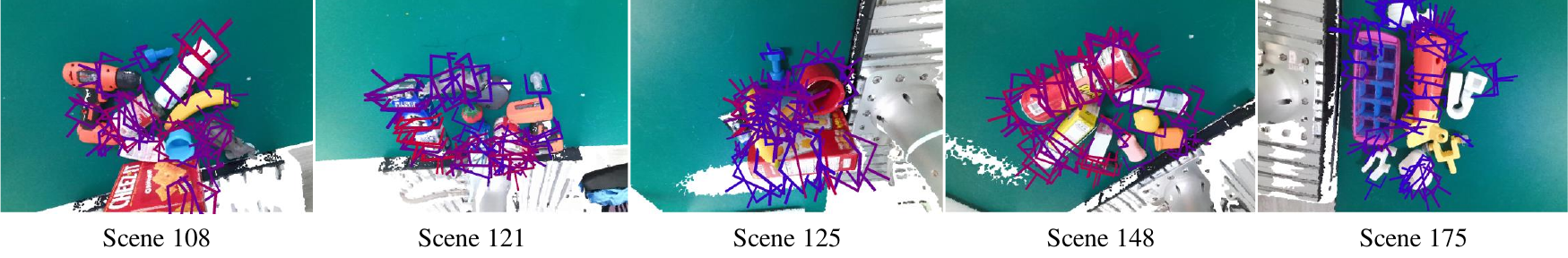}
    \caption{Qualitative results of GSNet. Top 50 grasps after grasp-NMS\cite{graspnet} are displayed.}
    \label{fig:result}
    \vspace{-0.2in}
\end{figure*}

%% file: charts/tab_fps_pvs.tex
\begin{table}[t]
    \centering\small
    \begin{tabular}{c|c|c}
        \hline
        Point Sampling & View Sampling & \textbf{AP}\\
        \hline
        random & graspable PVS & \footnotesize{46.17} \\
        FPS & graspable PVS & \footnotesize{51.15}\\
        graspable random & graspable PVS & \footnotesize{53.32}\\
        \hline
        graspable FPS & normal & \footnotesize{55.63}\\
        graspable FPS & top-1 score & \footnotesize{58.34}\\
        \hline
        graspable FPS & graspable PVS & \footnotesize{\textbf{59.70}}\\
        \hline
    \end{tabular}
    \caption{Comparison of different sampling methods.``top-1 score'' stands for selecting the view with the highest graspness score.}
    \label{tab:fps_pvs}
\end{table}

%% file: charts/tab_graspness_selection.tex
\begin{table}[t]
\begin{minipage}{0.48\linewidth}
    \centering\footnotesize
    \begin{tabular}{c|c}
        \hline
        Landscape & \textbf{AP} \\
        \hline
        object-level & 55.33 \\
        scene-level & \textbf{59.70}\\
        \hline
        \multicolumn{2}{c}{}
    \end{tabular}
    \caption{\ \ Landscape \ \ types.}
    \label{tab:point_graspness}
\end{minipage}
\begin{minipage}{0.48\linewidth}
    \centering\footnotesize
    \begin{tabular}{c|c}
        \hline
        View Graspness & \textbf{AP} \\
        \hline
        mean score & 50.62\\
        max score & 56.95 \\
        feasible ratio & \textbf{59.70}\\
        \hline
    \end{tabular}
    \caption{View graspness types.}
    \label{tab:view_graspness}
\end{minipage}
\vspace{-0.1in}
\end{table}

%% file: charts/tab_speed.tex
\begin{table}[t]
    \centering\small
    \begin{tabular}{p{2cm}<{\centering}|p{1.5cm}<{\centering}|p{1.5cm}<{\centering}|p{1.5cm}<{\centering}}
        \hline
        Camera & CGM & GOM & Total \\
        \hline
        RealSense & 0.08s & 0.02s & 0.10s \\
        Kinect & 0.10s & 0.02s & 0.12s \\
        \hline
    \end{tabular}
    \caption{Inference speed on GraspNet-1Billion. ``CGM'' is cascaded graspness model and ``GOM'' is grasp operation model.}
    \label{tab:speed}
    \vspace{-0.1in}
\end{table}

%% file: charts/tab_realgrasp_scene.tex
\begin{table}[t]
    \centering
    \scalebox{0.8}{
    \begin{tabular}{|c|c|c|c|}
        \hline
        IDs & \#Objects & \#Attempts & Success Rate \\
        \hline
        4, 10, 22, 32, 36, 57 & 6 & 6 & 100\% \\
        \hline
        2, 38, 58, 59, 61, 69 & 6 & 7 & 85.7\% \\
        \hline
        34, 37, 64, 66, 68, 72, 77 & 7 & 9 & 77.8\% \\
        \hline
        0, 2, 23, 29, 39, 56, 62 & 7 & 7 & 100\% \\
        \hline
        1, 10, 40, 41, 44, 48, 65, 69 & 8 & 8 & 100\% \\
        \hline
        3, 9, 10, 23, 33, 42, 63, 68 & 8 & 9 & 88.9\% \\
        \hline
        Total & 42 & 46 & 91.3\% \\
        \hline
    \end{tabular}}
    \caption{Results of cluttered scene grasping. \#Objects denotes the number of objects, and so does \#Attempts.}
    \label{tab:realgrasp_scene}
    \vspace{-0.1in}
\end{table}

%% file: 5_conclusion.tex
\section{Conclusion}
In this paper, we propose a novel geometrically based quality named graspness. A look-ahead searching method is adopted as our graspness measure and we statistically demonstrate its effectiveness and rationality. An end-to-end network is developed to incorporate graspness into grasp pose detection problem, wherein an independent model learns the graspable landscapes. We conduct extensive experiments and demonstrate the stability, generality, effectiveness and robustness of our graspness model. Large margin of improvements are witnessed for previous methods after equipping with our graspness model, and our final network sets a high record for both accuracy and speed.

\vspace{-0.1in}

\paragraph{Acknowledgement}
This work is supported in part by the National Key R\&D Program of China, No.2017YFA0700800, National Natural Science Foundation of China under Grants 61772332 and Shanghai Qi Zhi Institute, SHEITC(2018-RGZN-02046).

%% file: 6_appendix.tex
\clearpage
\appendix

\section{Video Demo and Library}
A video demo is attached in the supplementary files\footnote{\url{https://openaccess.thecvf.com/content/ICCV2021/supplemental/Wang_Graspness_Discovery_in_ICCV_2021_supplemental.zip}} for real grasping using the results predicted by GSNet which is trained on GraspNet-1Billion. Watch the video ``demo.mp4'' for more details. Notably, some objects (chain, mesh bag with marbles, slipper, etc.) in the demo are not collected from GraspNet-1Billion~\cite{graspnet} and our model shows robustness on novel objects.

GSNet library has been upgraded and integrated into AnyGrasp~\cite{fang2023anygrasp}. See the project website\footnote{\url{https://graspnet.net/anygrasp}} for details and more videos.

\section{Grasping Experiment Configuration}
\input{charts/fig_supp_exp_setting}
Fig.~\ref{fig:supp_exp_setting} shows the configuration of our grasping experiments. A grasp with a high score output by GSNet is chosen and sent to the robotic arm. The program attempts to grab one object each time, and repeat execution until all the objects are cleaned from the table.

\section{Robotic Experiments with Baselines}
\input{charts/tab_supp_baselines}
We compare our methods with other baselines~\cite{graspnet, pointnetgpd, gpd} in real experiments. Objects are divided into three sets, each containing 10 objects from \cite{graspnet}. These methods are used to remove all the objects in the workspace with single-view point clouds as input. Four repeated experiments are conducted for each object set. Tab.~\ref{tab:supp_baselines} shows the results of different methods, where success rate is defined in Sec. 5.6. GSNet outperforms other methods on all three sets.

\section{Details of Grasp Operation Model}
\input{charts/fig_supp_graspness}

Grasp operation model (GOM) in GSNet is designed based on OperationNet in GraspNet baseline model~\cite{graspnet} with several improvements. The main differences between the two components are listed as follows.

\paragraph{Simplified Cylinder Representation} In~\cite{graspnet}, points are cropped and transformed into a cylinder region for each depth bin, which leads to multiple groups with repeated points on one grasp proposal. GOM replaces them with a single cylinder region, where the height is determined by the maximum depth (0.04m) used in our experiments. Depth classification is moved to final output accordingly.

\paragraph{Scaled Point Coordinates} In~\cite{graspnet}, points are transformed without scaling. Since all groups shared the same gripper coordinate frame and the transformed coordinates are relative small in abusolute value ($<=$ 0.05m), we scaled the points with the cylinder radius (0.05m). The width prediction is modified accordingly.

\paragraph{Shared Point Features} For each depth bin on one grasp point, OperationNet in~\cite{graspnet} directly samples 64 points with only xyz coordinates from the original input (about 20k points). GOM samples 16 points from the seeds (about 1k points). The xyz coordinates are then concatenated with point features output by cascaded grasp model. This modification helps reduce computing overhead of point sampling.

\paragraph{Output Parameters} \cite{graspnet} output grasp scores, in-plane rotation angles and gripper widths for each depth bin, and choose the parameters with the highest angle classification scores. In GOM, grasp scores and gripper widths are predicted for each (in-plane rotation)-(approach depth) combination and output the parameters of the combination with the highest grasp score. In addition, output grasp scores and gripper widths are replaced with relative value from 0 to 1.

\section{Visualization of Point-wise Graspness}
\input{charts/tab_supp_tsne_id}
Point-wise graspness predicted by GSNet is visualized in Fig.~\ref{fig:supp_graspness}. Regions with higher graspness are annotated with brighter colors. We can see that graspness is not only decided by the object itself, but also influenced by its position. Most of the low-graspness areas are caused by collision with tables, which have higher graspness in single object. Graspness of an object is also influenced by its neighbours. For example, the knife on the banana (Fig.~\ref{fig:supp_graspness}d) cut off the contiguous graspness of the latter. The object size is also an important factor. In Fig.~\ref{fig:supp_graspness}c, the box has no areas with high graspness although the shape is relative simple. That is because there are few areas for a gripper with width up to 0.1m to grasp on when the box are lying on the table.

\section{t-SNE Visualization of Point Features}

\input{charts/fig_supp_tsne}
We visualize the point features output by GSNet.
Fig.~\ref{fig:supp_tsne} shows the t-SNE visualization of point features in different test setting, where points are obtained from GraspNet-1Billion dataset and feature vectors are output by GSNet. Tab.~\ref{tab:supp_tsne} details the experimental settings. The points with high graspness are labeled as positive samples, and other points are labeled as negative samples. We can see that graspable points are quite distinguishable from others, which demonstrates the generality of graspness model across different settings.

%% file: charts/fig_supp_exp_setting.tex
\begin{figure}[h]
    \centering
    \includegraphics[width=\linewidth]{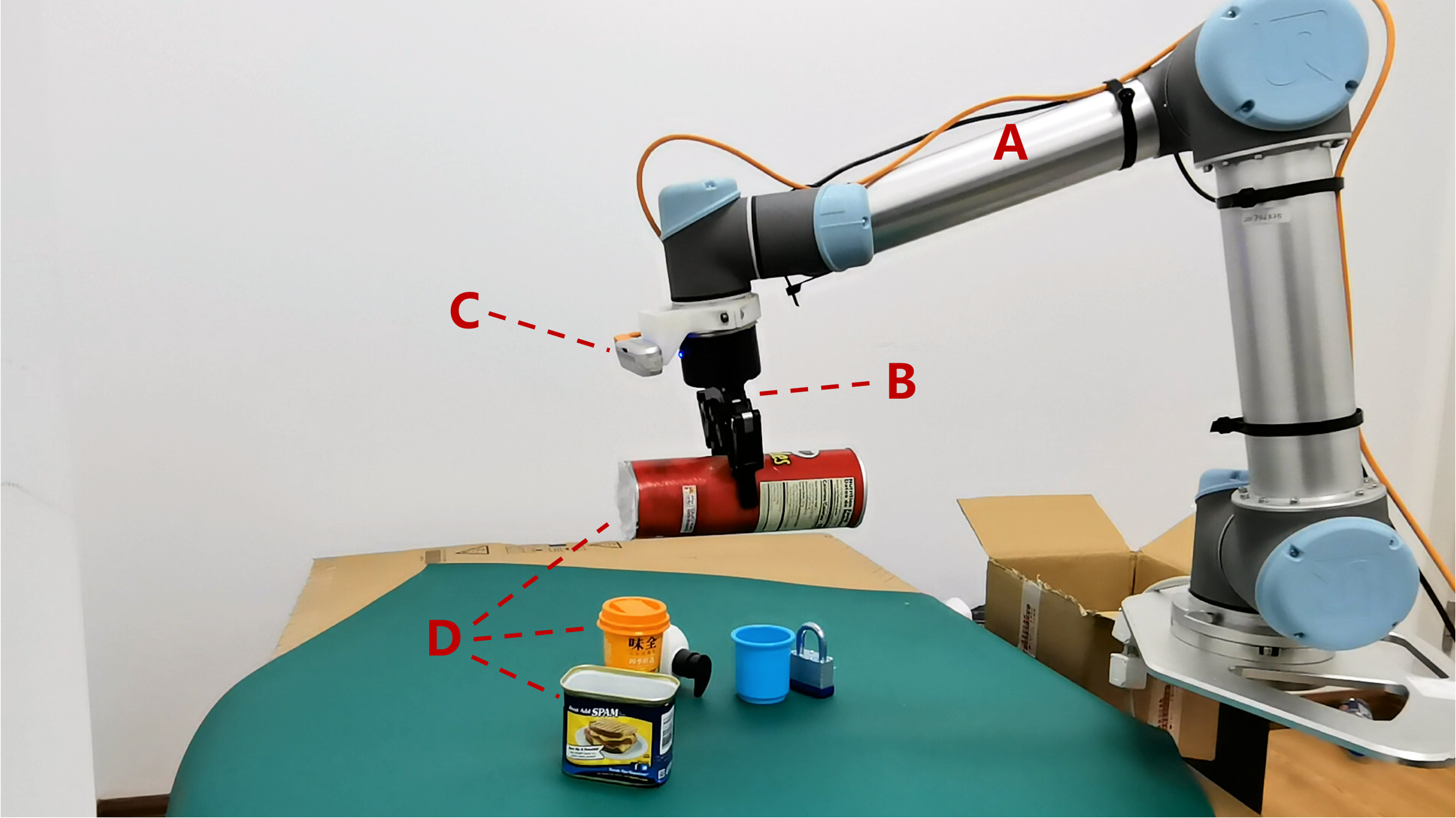}
    \caption{Configuration of real grasping experiments. A: UR-5 robotic arm. B: Robotiq two-finger gripper. C: RealSense D435 camera. D: object models from GraspNet-1Billion dataset.}
    \label{fig:supp_exp_setting}
\end{figure}

%% file: charts/tab_supp_baselines.tex
\begin{table}[h]
    \centering
    \scalebox{0.8}{
    \begin{tabular}{|c|c|c|c|}
        \hline
        \multirow{2}{*}{Method} & \multicolumn{3}{c|}{Success Rate}\\
        \cline{2-4}
         & Set A & Set B & Set C \\
        \hline
        GPD~\cite{gpd} & 71.43\% & 74.07\% & 67.80\%\\
        \hline
        PointNetGPD~\cite{pointnetgpd} & 76.92\% & 78.43\% & 70.18\%\\
        \hline
        Fang~\etal~\cite{graspnet} & 81.63\% & 80.00\% & 74.07\%\\
        \hline
        Ours & \textbf{88.89\%} & \textbf{90.91\%} & \textbf{85.11\%}\\
        \hline
    \end{tabular}}
    \caption{Success rate on real robot experiments.}
    \label{tab:supp_baselines}
\end{table}

%% file: charts/fig_supp_graspness.tex
\begin{figure*}[t]
    \centering
    
    \subfigure[Scene 103]
    {
        \centering
        \includegraphics[width=0.23\linewidth]{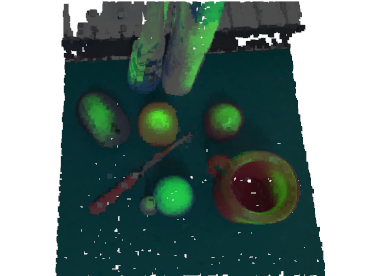}
    }
    \subfigure[Scene 105]
    {
        \centering
        \includegraphics[width=0.23\linewidth]{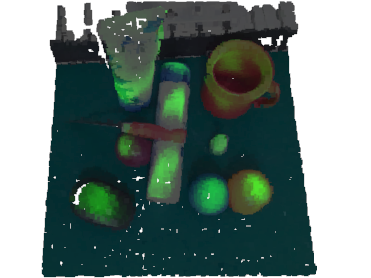}
    }
    \subfigure[Scene 106]
    {
        \centering
        \includegraphics[width=0.23\linewidth]{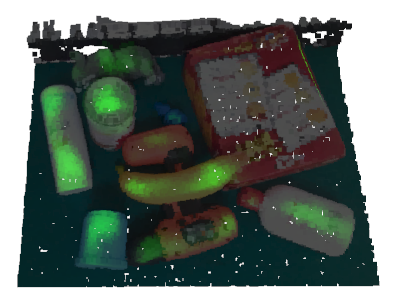}
    }
    \subfigure[Scene 113]
    {
        \centering
        \includegraphics[width=0.23\linewidth]{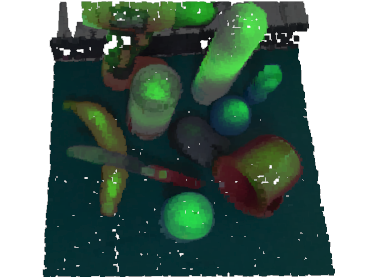}
    }
    
    \subfigure[Scene 118]
    {
        \centering
        \includegraphics[width=0.23\linewidth]{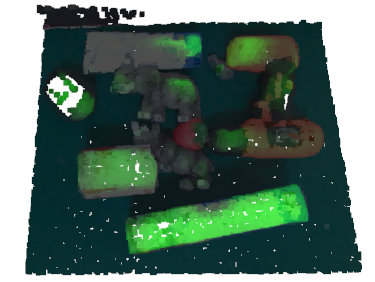}
    }
    \subfigure[Scene 146]
    {
        \centering
        \includegraphics[width=0.23\linewidth]{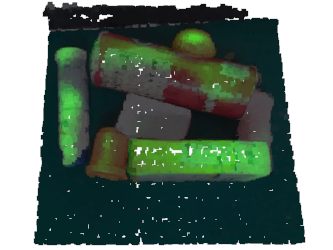}
    }
    \subfigure[Scene 150]
    {
        \centering
        \includegraphics[width=0.23\linewidth]{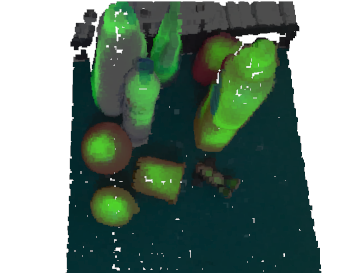}
    }
    \subfigure[Scene 183]
    {
        \centering
        \includegraphics[width=0.23\linewidth]{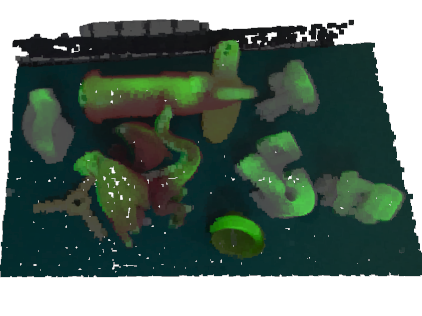}
    }
    
    \caption{Visualization of point-wise graspness predicted by GSNet. Regions with higher graspness are annotated with brighter colors.}
    \label{fig:supp_graspness}
\end{figure*}

%% file: charts/tab_supp_tsne_id.tex
\begin{table*}[!h]
    \centering\small\scalebox{0.96}{
    \begin{tabular}{c||c|ccc||c|ccc||c|cc}
        \hline
        & \multicolumn{4}{c||}{Object Variation} & \multicolumn{4}{c||}{Viewpoint Variation} & \multicolumn{3}{c}{Camera Variation}\\
        \cline{2-12}
        & Train & \multicolumn{3}{c||}{Test} & Train & \multicolumn{3}{c||}{Test} & Train & \multicolumn{2}{c}{Test} \\
        \hline
        Scene & 0-99 & 100-129 & 130-159 & 160-189 & 0-99 & 100-129 & 100-129 & 100-129 & 0-99 & 100-129 & 100-129 \\
        \hline
        View & 0-255 & 0-255 & 0-255 & 0-255 & 0-127 & 0-127 & 128-191 & 192-255 & 0-255 & 0-255 & 0-255 \\
        \hline
        Camera & Kinect & Kinect & Kinect & Kinect & Kinect & Kinect & Kinect & Kinect & Kinect & Kinect & Realsense \\
        \hline
        Serial Number & A1 & A2 & A3 & A4 & B1 & B2 & B3 & B4 & C1 & C2 & C3\\
        \hline
    \end{tabular}}
    \caption{Serial number on different test setting. Each number corresponds to a t-SNE visualization result in Fig.~\ref{fig:supp_tsne}.}
    \label{tab:supp_tsne}
\end{table*}

%% file: charts/fig_supp_tsne.tex
\begin{figure*}[t]
    \centering
    
    \subfigure[A1]
    {
        \centering
        \fbox{\includegraphics[width=0.22\linewidth]{figures/fig_supp_tsne/A1.jpg}}
    }
    \subfigure[A2]
    {
        \centering
        \fbox{\includegraphics[width=0.22\linewidth]{figures/fig_supp_tsne/A2.jpg}}
    }
    \subfigure[A3]
    {
        \centering
        \fbox{\includegraphics[width=0.22\linewidth]{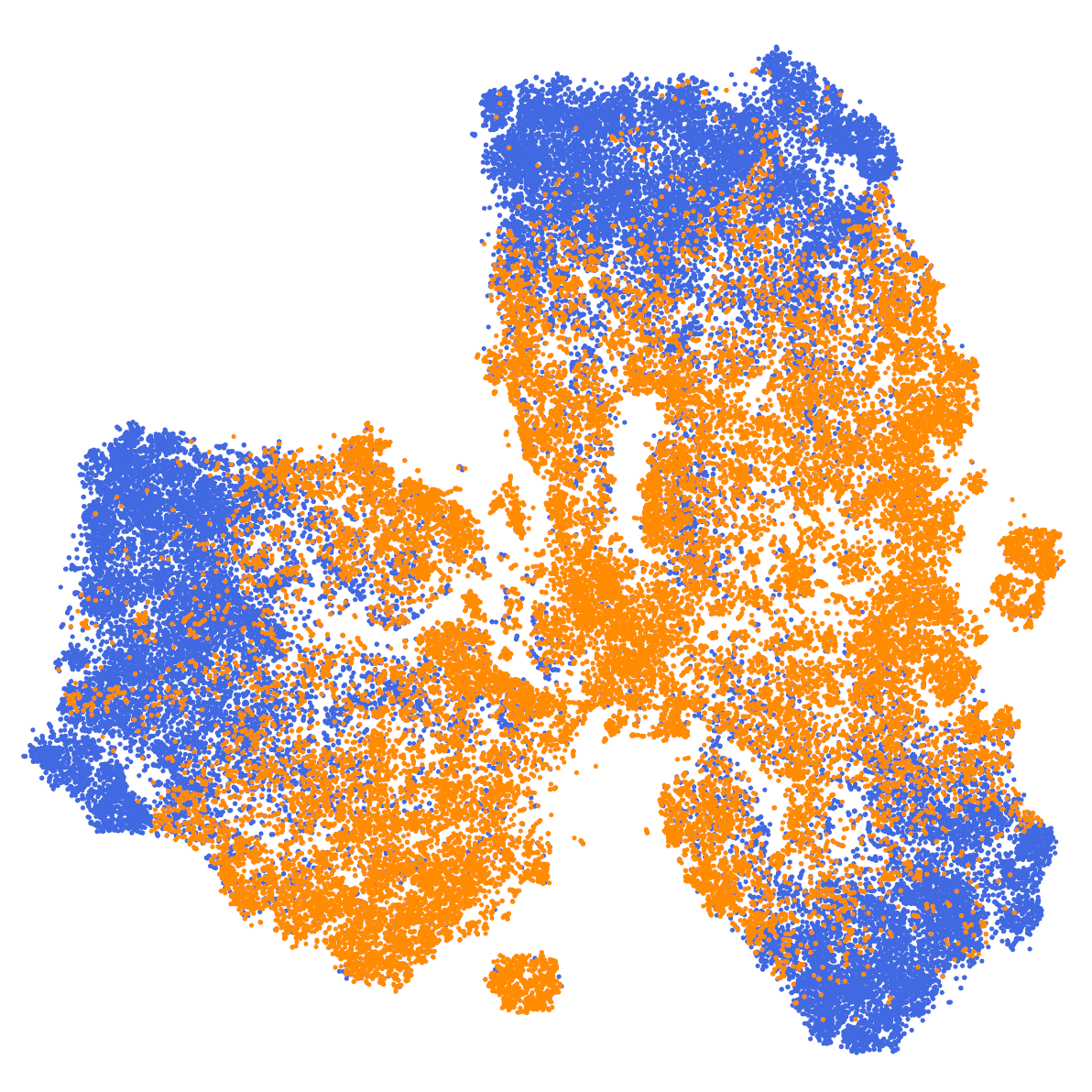}}
    }
    \subfigure[A4]
    {
        \centering
        \fbox{\includegraphics[width=0.22\linewidth]{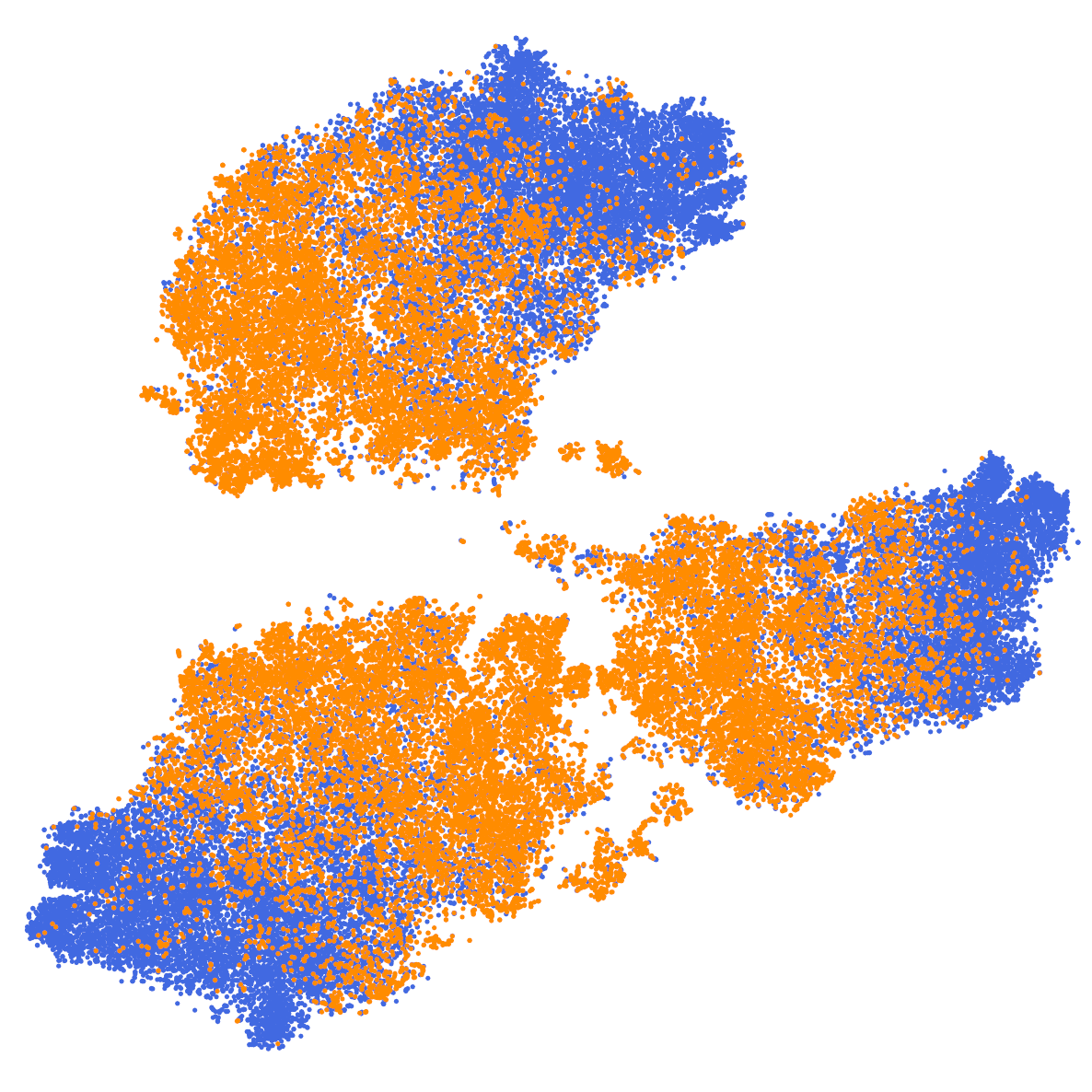}}
    }
    
    \subfigure[B1]
    {
        \centering
        \fbox{\includegraphics[width=0.22\linewidth]{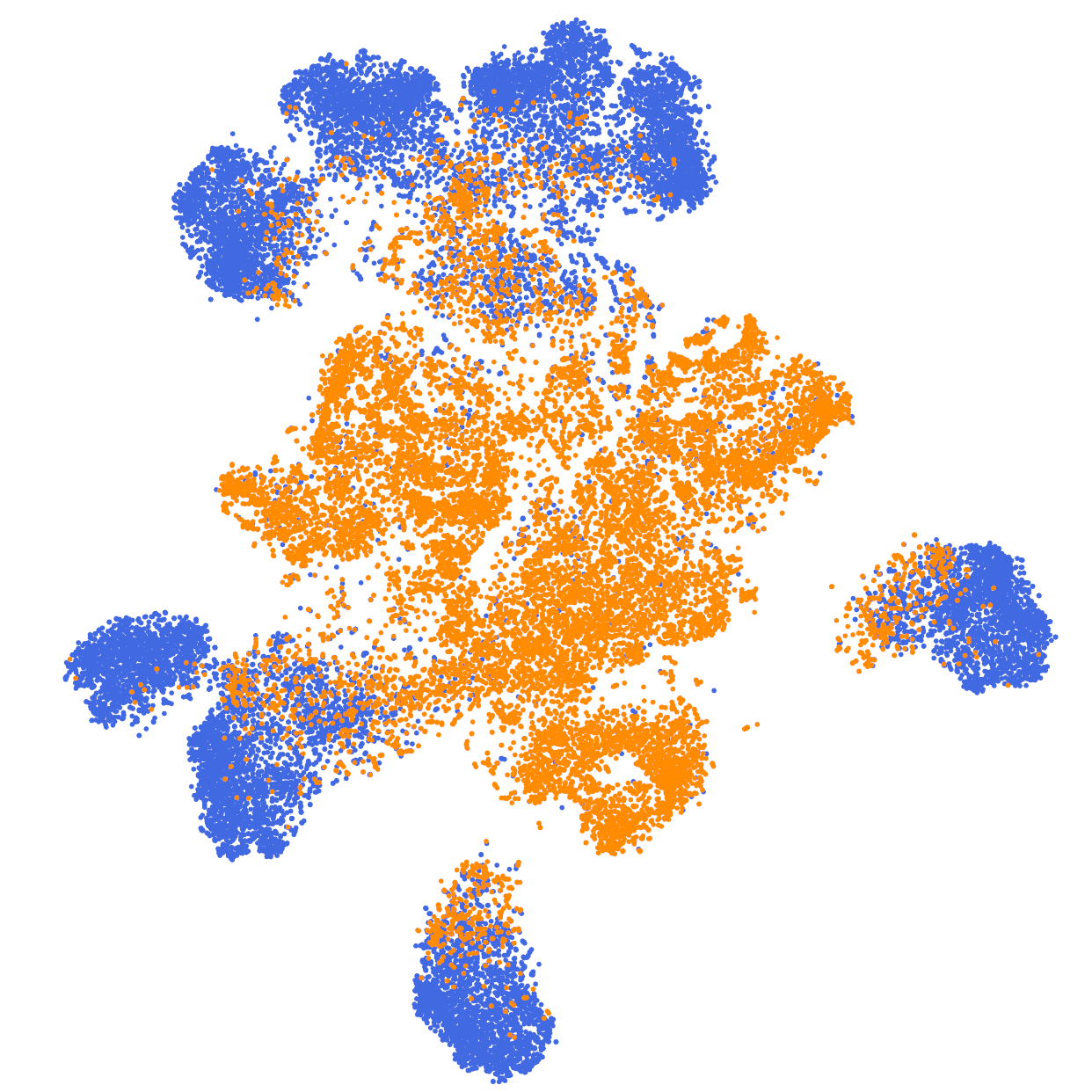}}
    }
    \subfigure[B2]
    {
        \centering
        \fbox{\includegraphics[width=0.22\linewidth]{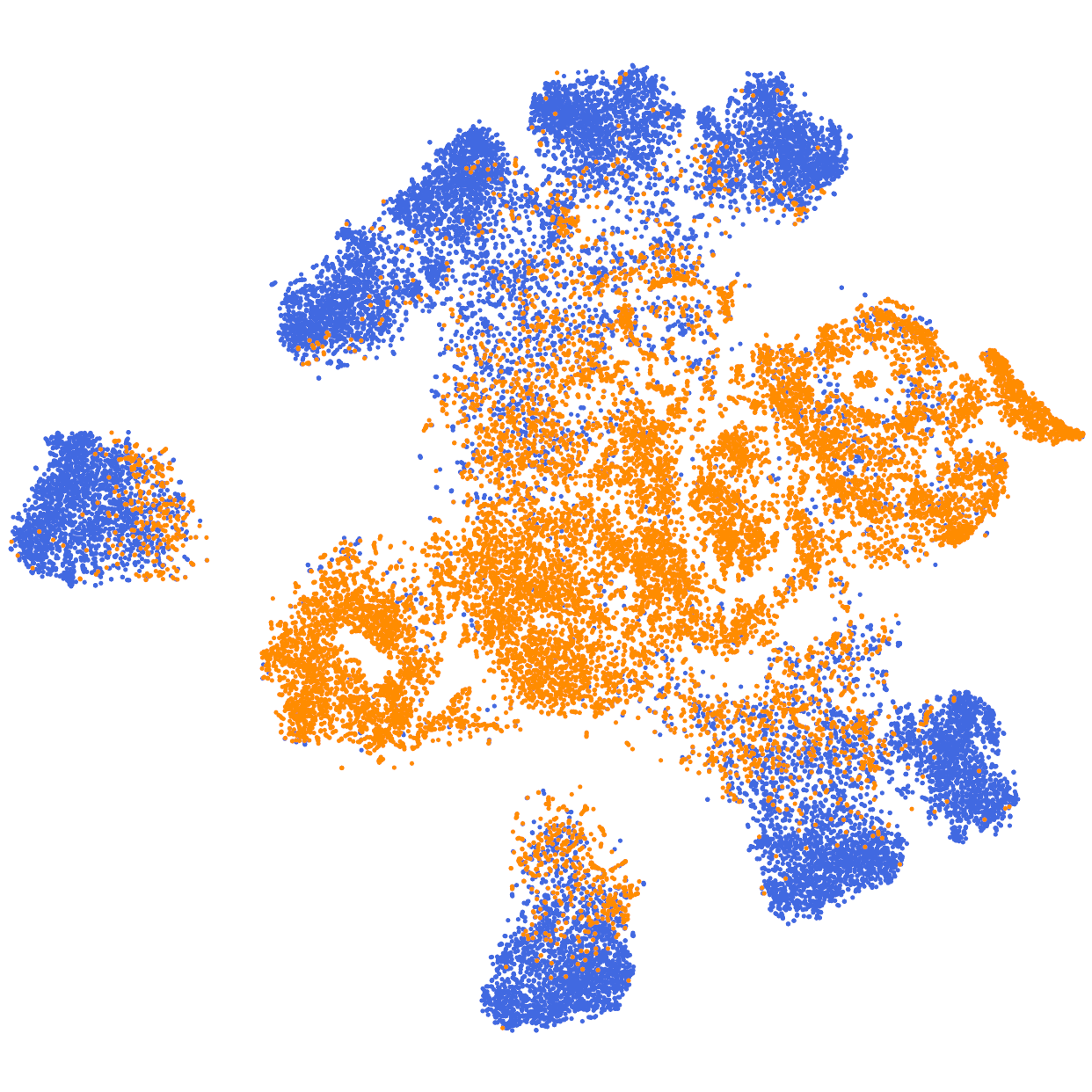}}
    }
    \subfigure[B3]
    {
        \centering
        \fbox{\includegraphics[width=0.22\linewidth]{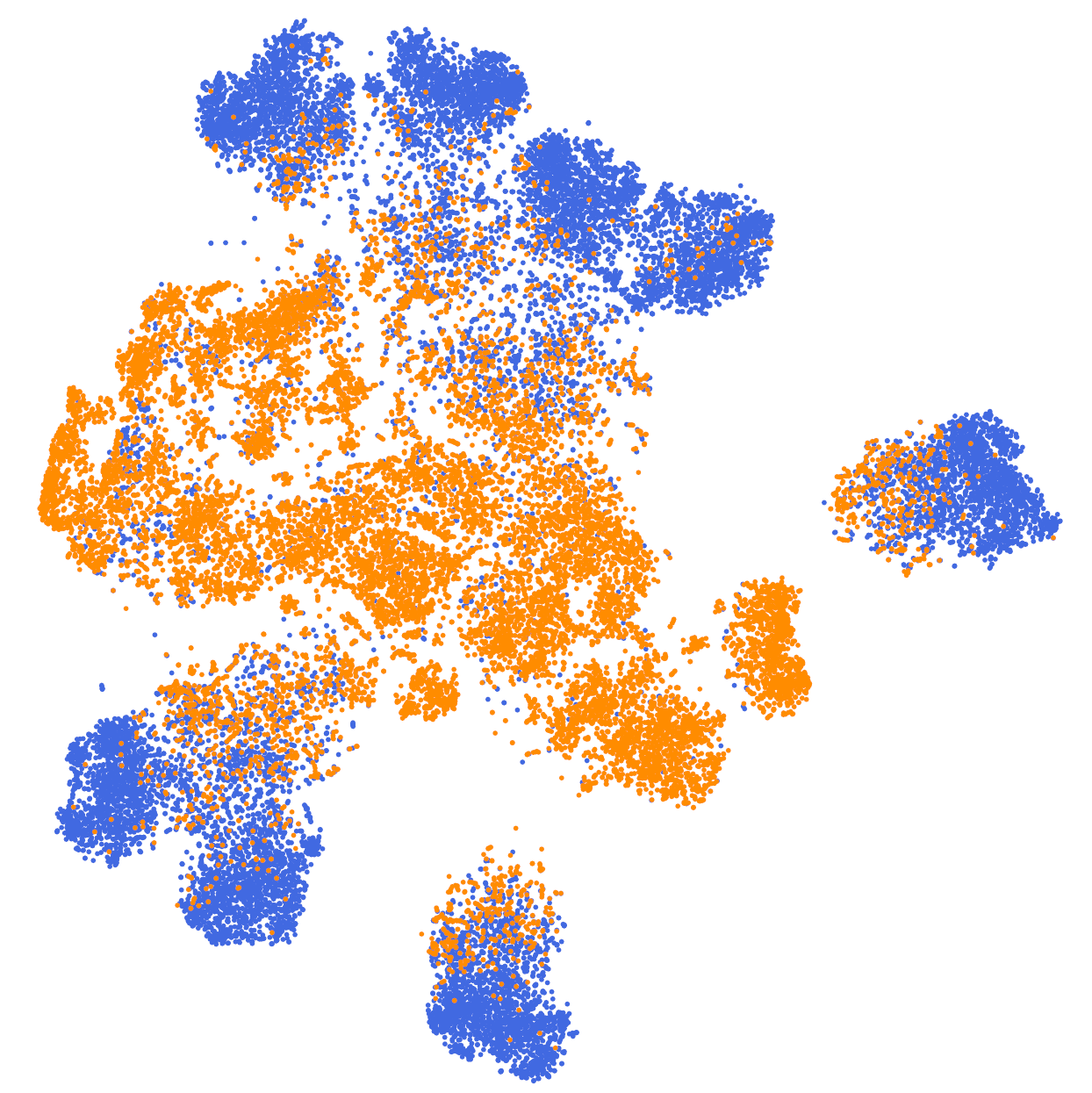}}
    }
    \subfigure[B4]
    {
        \centering
        \fbox{\includegraphics[width=0.22\linewidth]{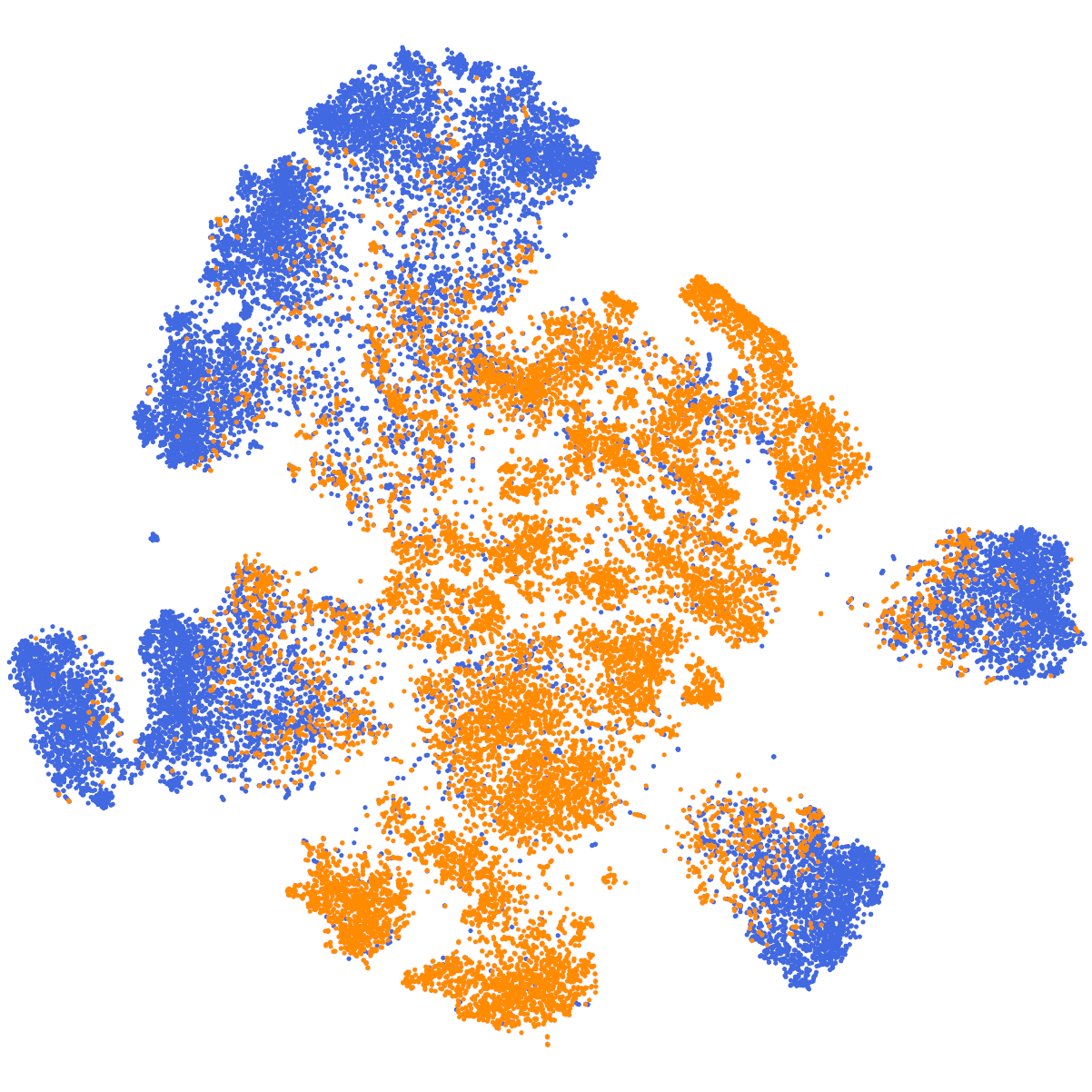}}
    }
    
    \subfigure[C1]
    {
        \centering
        \fbox{\includegraphics[width=0.22\linewidth]{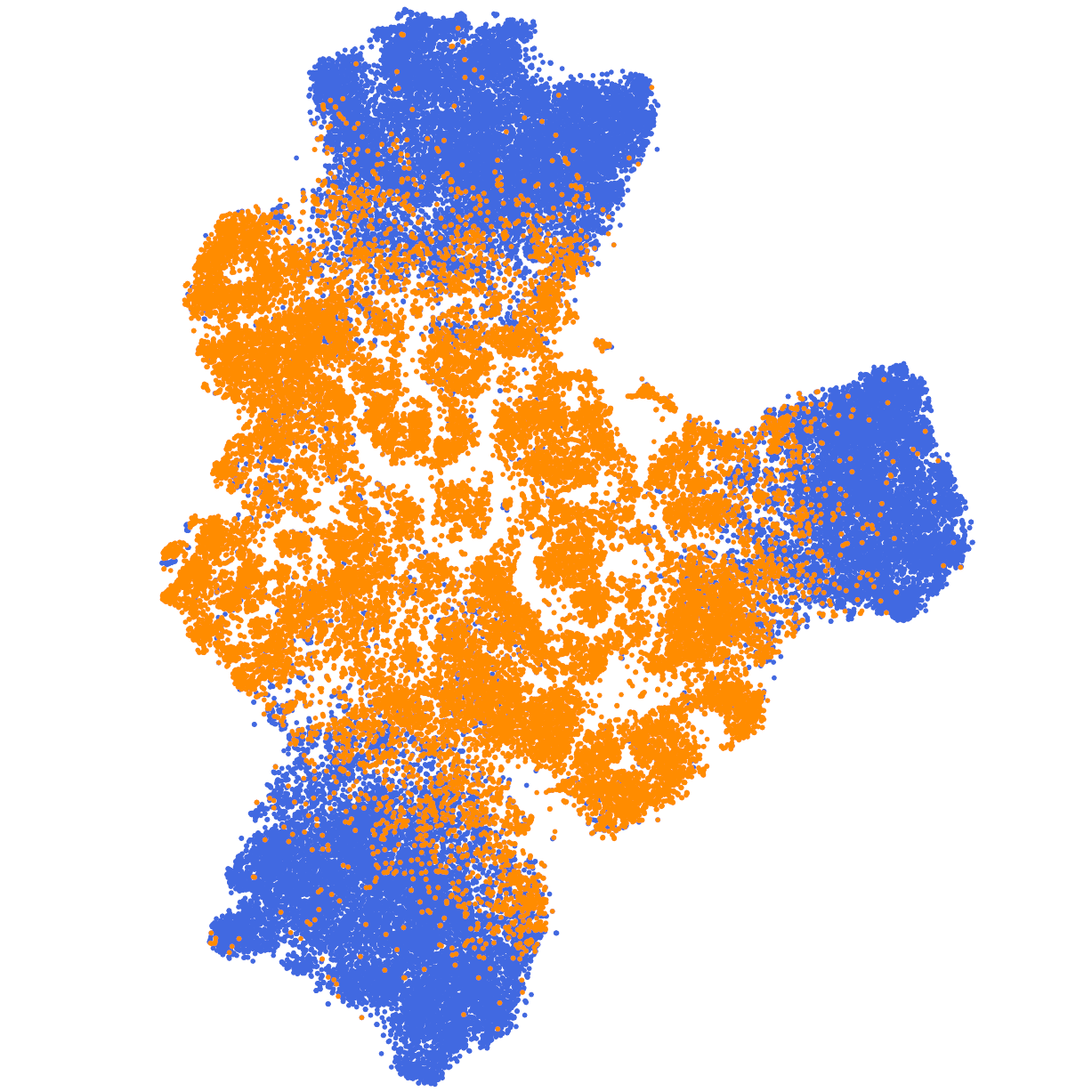}}
    }
    \subfigure[C2]
    {
        \centering
        \fbox{\includegraphics[width=0.22\linewidth]{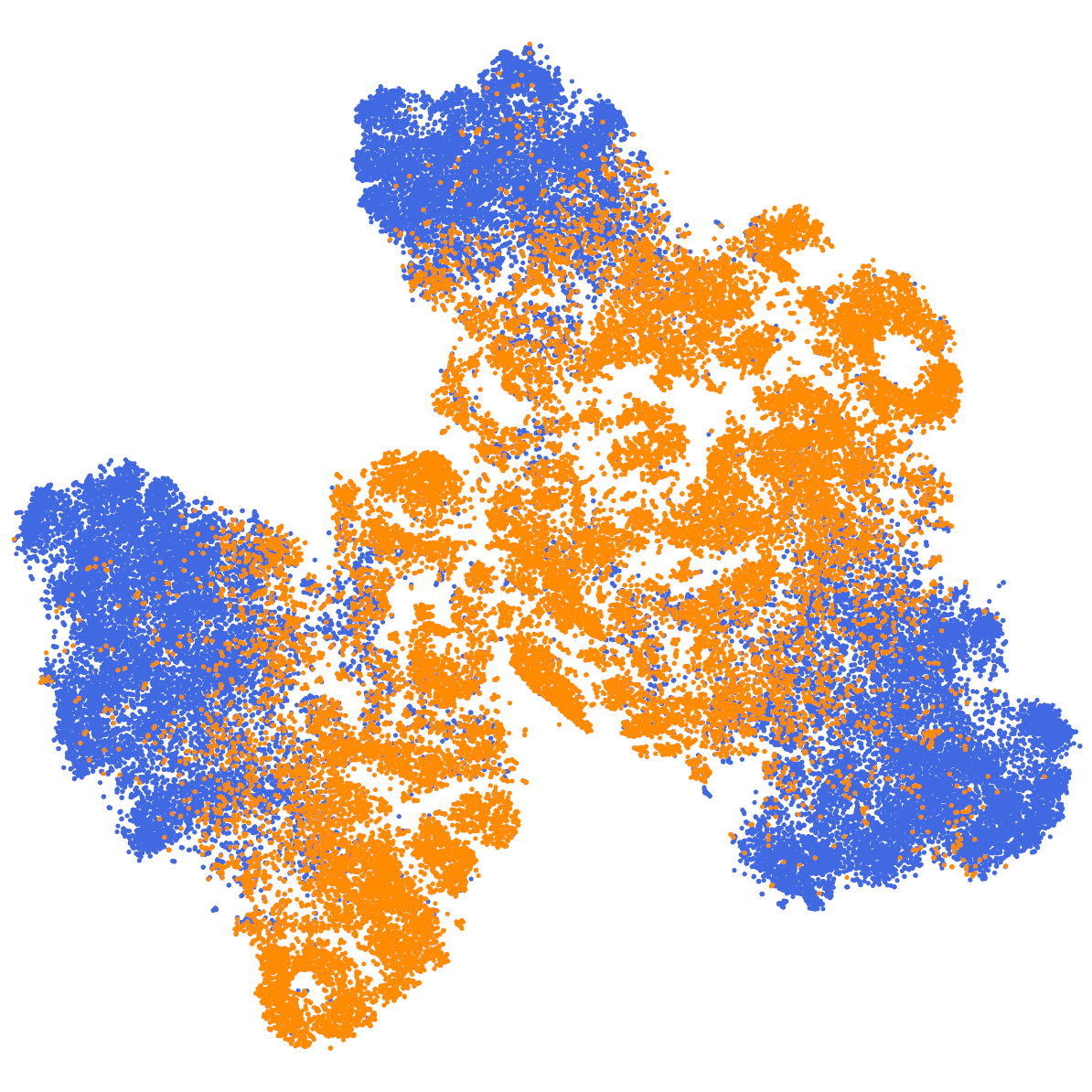}}
    }
    \subfigure[C3]
    {
        \centering
        \fbox{\includegraphics[width=0.22\linewidth]{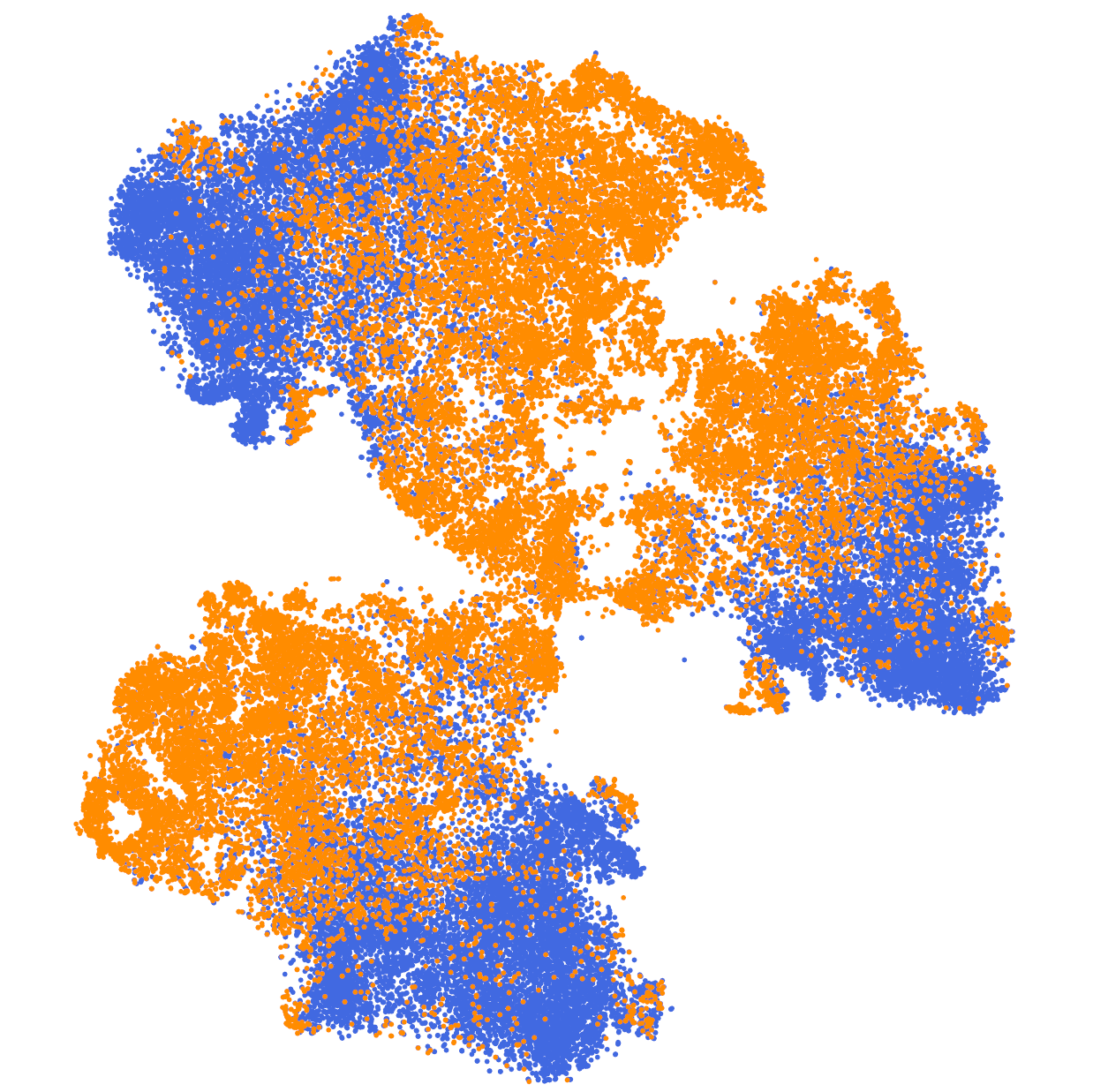}}
    }
    
    \caption{t-SNE visualization of point features on different test setting. The three rows show the results of object/viewpoint/camera variation respecctively. Orange points denote the samples with high graspness, and blue points denote the samples with low graspness. The setting details are listed in Tab.~\ref{tab:supp_tsne}.}
    \label{fig:supp_tsne}
\end{figure*}